\documentclass[letterpaper,10pt,conference]{ieeeconf}

\IEEEoverridecommandlockouts
\usepackage{graphicx}
\usepackage{booktabs}
\usepackage{amsmath,amssymb}
\usepackage{multirow}
\usepackage{tabularx}
\usepackage{array}
\usepackage{float}
\usepackage{algorithm}
\usepackage[breaklinks,colorlinks=true,linkcolor=blue,citecolor=blue,urlcolor=blue]{hyperref}
\usepackage{placeins}
\usepackage{cite}
\usepackage{dblfloatfix}
\usepackage{cuted}
\usepackage{capt-of}
\usepackage{needspace}


\title{\LARGE \bf
Splat2Real: Novel-view Scaling for Physical AI\\
with 3D Gaussian Splatting}

\author{Hansol Lim$^{1,2}$ and Jongseong Brad Choi$^{1,2}$%
\thanks{$^{1}$Department of Mechanical Engineering, State University of New York, Korea, 21985, Republic of Korea.}%
\thanks{$^{2}$Department of Mechanical Engineering, State University of New York, Stony Brook, Stony Brook, NY 11794, USA.}%
\thanks{E-mails: hansol.lim@stonybrook.edu; jongseong.choi@stonybrook.edu.}%
\thanks{This work is supported by the National Research Foundation of Korea (NRF) grant funded by the Korea government (MSIT) (Grant No. RS-2022NR067080 and RS-2025-05515607).}%
\thanks{Corresponding author: Jongseong Brad Choi.}}

\begin{document}

\maketitle
\thispagestyle{empty}
\pagestyle{empty}

\begin{abstract}
Physical AI faces viewpoint shift between training and deployment, and \emph{novel-view robustness} is essential for monocular RGB-to-3D perception. We cast Real2Render2Real monocular depth pretraining as imitation-learning--style supervision from a digital twin oracle: a student depth network imitates expert metric depth/visibility rendered from a scene mesh, while 3DGS supplies scalable novel-view observations. We present \emph{Splat2Real}, centered on \textbf{novel-view scaling}: performance depends more on \emph{which} views are added than on raw view count. We introduce \textbf{CN-Coverage}, a coverage+novelty curriculum that greedily selects views by geometry gain and an extrapolation penalty, plus a quality-aware \textbf{guardrail fallback} for low-reliability teachers. Across 20 TUM RGB-D sequences with step-matched budgets ($N=0$ to $2000$ additional rendered views, with $N_{\text{unique}}\leq500$ and resampling for larger budgets), naive scaling is unstable; CN-Coverage mitigates worst-case regressions relative to Robot/Coverage policies, and GOL-Gated CN-Coverage provides the strongest medium/high-budget stability with the lowest high-novelty tail error. Downstream control-proxy results versus $N$ provide embodied-relevance evidence by shifting safety/progress trade-offs under viewpoint shift.
\end{abstract}

\section{Introduction}
Physical AI agents must make geometry-aware decisions under viewpoint shift: deployment cameras observe poses that are sparse or absent in training data. In many stacks, imitation learning (IL) policies depend on robust perception; here we study IL-style supervision for the perception module itself. We do not train a control policy with IL; we train a monocular depth model to imitate a digital-twin oracle that provides depth/visibility supervision.

A practical route is to decouple appearance and geometry supervision. 3D Gaussian Splatting (3DGS) provides fast novel-view rendering from real captures, while simulator-style mesh rendering provides aligned dense depth labels and visibility masks. In this Real2Render2Real setting, the main question is \emph{how} to scale views: adding many poorly chosen views can hurt transfer stability.

We therefore make \textbf{novel-view scaling} the central object of study. We introduce \textbf{Splat2Real}, where 3DGS provides scalable observation rendering and mesh rendering provides metric supervision. Our main methodological contribution is \textbf{CN-Coverage}, a coverage+novelty curriculum that greedily selects viewpoints by geometry gain and extrapolation penalty. Quality-gated/composited observation mixing is a secondary guardrail when teacher reliability is low.

\textbf{Contributions.}
\begin{itemize}
\item We present \textbf{Splat2Real}, framing novel-view scaling as imitation-learning--style supervision for monocular perception: a student imitates a digital twin oracle (mesh depth/visibility) while 3DGS supplies rendered observations.
\item We introduce \textbf{CN-Coverage}, a coverage+novelty scaling policy motivated by pose-distribution shift and submodular-style coverage with diminishing returns, and pair it with \textbf{Gaussian Observation Layers (GOL)} guardrails for reliability-aware fallback.
\item We provide a 20-sequence step-matched study with additional rendered-view budgets from $N=0$ to $N=2000$ ($N_{\text{unique}}\leq500$), including scaling stability, high-novelty tail robustness, and downstream control-proxy trade-offs.
\end{itemize}

\section{Related Work}
\paragraph{Monocular depth and rendering-based supervision.}
Monocular depth has progressed rapidly with transformer/hybrid architectures such as DPT~\cite{ranftl2021dpt}, but robustness remains supervision-limited. Neural rendering for viewpoint augmentation has been explored for depth learning (e.g., NeRFmentation~\cite{feldmann2024nerfmentation}) and for splatting-depth integration (DepthSplat~\cite{xu2024depthsplat}). Our focus is not architecture novelty; it is how viewpoint scaling policies affect Sim2Real transfer.

\paragraph{Sim2Real adaptation for perception.}
Domain randomization~\cite{tobin2017domain} and adversarial adaptation~\cite{ganin2016domain} are standard tools for closing visual gaps, but gains can be inconsistent under large deployment viewpoint shift. We study an orthogonal axis: scaling viewpoint support with high-throughput learned rendering while keeping metric labels fixed.

\paragraph{Imitation learning and teacher--student supervision.}
Imitation learning (IL) and behavior cloning treat learning as matching expert outputs from supervised data~\cite{hussein2017imitation,osa2018algorithmic}. Teacher--student supervision with privileged targets is a common mechanism for transferring richer supervision into deployable models~\cite{hinton2015distilling}. Our use of IL terminology is module-specific: we imitate oracle geometry outputs for monocular perception pretraining, not expert actions or control trajectories.

\paragraph{Novel-view scaling and view selection.}
Rendering-based viewpoint scaling is a long-standing idea (Render for CNN~\cite{su2015render}). Recent depth work uses learned novel-view synthesis for augmentation~\cite{feldmann2024nerfmentation}. Safety-aware augmentation methods (e.g., AugMix~\cite{hendrycks2020augmix}) show that augmentation quality can matter more than augmentation count. Our CN-Coverage policy connects this to explicit pose/coverage objectives. We use a coverage-greedy objective inspired by submodular optimization and robotics view-planning/NBV literature~\cite{krause2014submodular,bircher2016receding}. Unlike learned NBV policies designed for online reconstruction quality, our setting is offline training-data synthesis for depth robustness under deployment viewpoint shift.

\paragraph{3DGS in embodied simulation and NBV scope.}
3DGS~\cite{kerbl2023gs} is widely used in digital twins and embodied Sim2Real pipelines~\cite{qureshi2025splatsim,li2024robogsim,wu2024rlgsbridge,chhablani2025embodiedsplat}. These efforts focus on policy transfer, active mapping, or reconstruction quality. We focus on offline training-view selection for depth robustness under viewpoint shift, and therefore compare lightweight, reproducible coverage/novelty samplers rather than full learned NBV controllers.

\paragraph{Datasets.}
For complete reproducibility of this pipeline, we evaluate on a 20-sequence TUM RGB-D benchmark~\cite{sturm2012benchmark} with per-sequence statistics and fixed train/val/test splits.

\section{Method}
\begin{figure*}[!htbp]
    \centering
    \includegraphics[width=\textwidth]{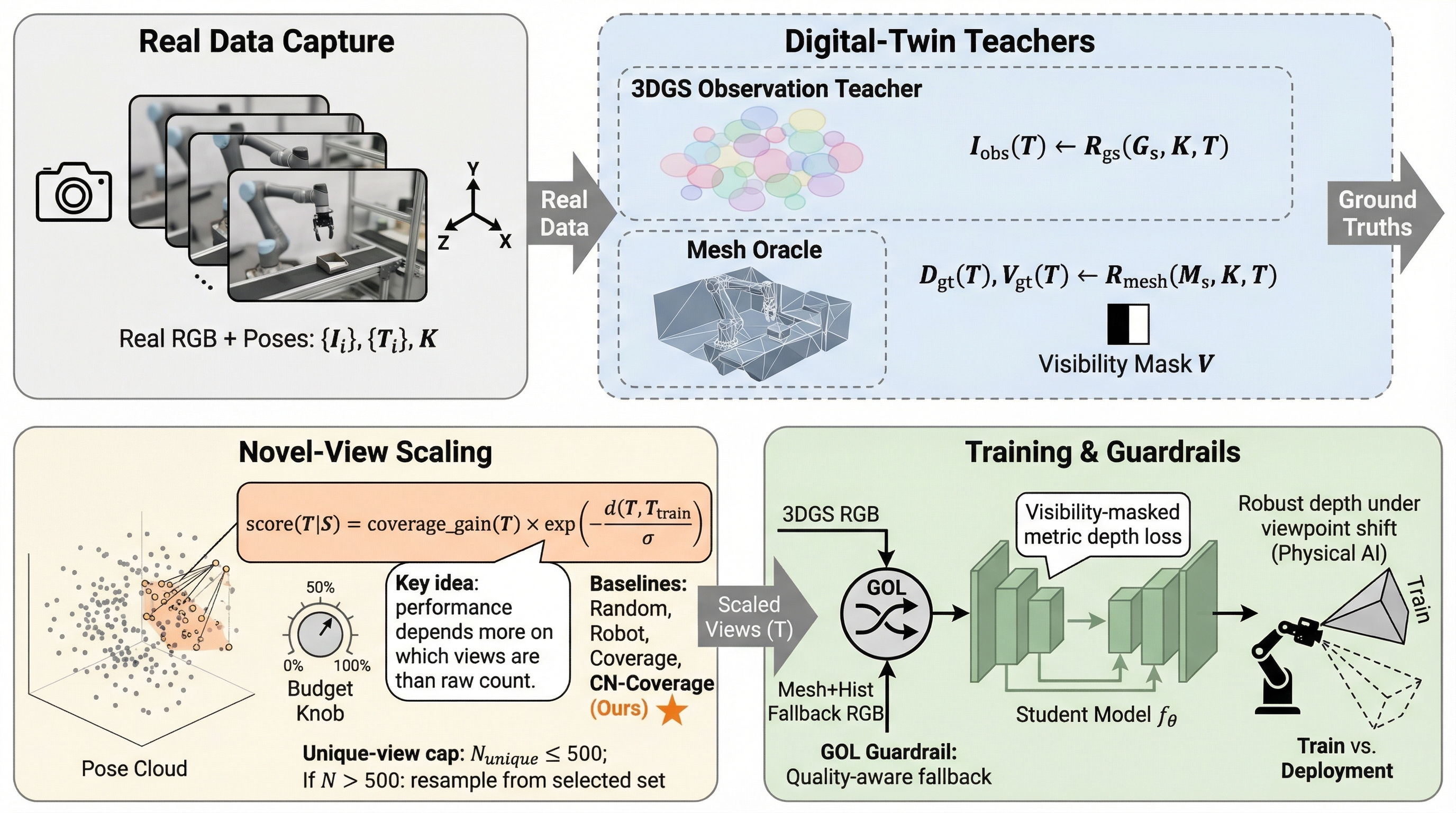}
    \caption{Splat2Real pipeline. Real captures build a 3DGS observation teacher for high-throughput novel-view RGB rendering; simulator-style mesh rendering provides aligned metric oracle labels. A scaling policy (Random/Robot/Coverage/CN-Coverage) selects viewpoint budget $N$. Gated/composited fallback is used as a secondary safety layer.}
    \label{fig:method}
\end{figure*}

Given scene $s$ with intrinsics $\mathbf{K}$ and poses $\{\mathbf{T}_i\}$, we train monocular depth model $f_\theta$ from tuples
\begin{align}
I_{\mathrm{obs}}(\mathbf{T}) &\leftarrow \mathcal{R}_{\mathrm{gs}}(G_s,\mathbf{K},\mathbf{T}) \;\text{or fallback blend},\\
D_{\mathrm{gt}}(\mathbf{T}),V_{\mathrm{gt}}(\mathbf{T}) &\leftarrow \mathcal{R}_{\mathrm{mesh}}(M_s,\mathbf{K},\mathbf{T}).
\end{align}
Depth supervision is metric and visibility-masked.

\paragraph{IL view (perception-only).}
We treat training as imitation-learning--style teacher--student supervision for perception. The digital-twin mesh renderer is the expert/oracle that provides $(D_{\mathrm{gt}},V_{\mathrm{gt}})$ at each pose $\mathbf{T}$, and the student $f_\theta$ maps RGB observations to depth. The imitation target is visibility-masked metric depth matching under our training loss, i.e., behavior cloning for perception outputs rather than action imitation.

\subsection{Novel-view scaling objective}
\paragraph{Pose-distribution shift lens.}
Let $p_{\text{train}}(\mathbf{T})$ be the training pose distribution induced by a sampling policy and $p_{\text{robot}}(\mathbf{T})$ the deployment distribution. Increasing $N$ helps only when it reduces mismatch between these distributions. Random oversampling can increase mass on low-value or extrapolative poses, improving image count but not deployment-aligned coverage.

\paragraph{Coverage lens (weighted submodularity).}
For selected pose set $S$, define visible-surface coverage
\begin{equation}
F(S)=\left|\bigcup_{\mathbf{T}\in S}V(\mathbf{T})\right|.
\end{equation}
We implement $V(\mathbf{T})$ as the set of occupied voxels obtained by rendering mesh depth at pose $\mathbf{T}$ (z-buffer visibility), backprojecting valid depth pixels, and voxelizing world points. Let voxel size be $\nu$ and downsample stride be $r$; both are fixed per experiment. Coverage is normalized per scene by the union over all sampled modes/budgets to produce comparable coverage fractions; exact parameters are listed in Appendix~B.

Coverage has diminishing returns; greedy addition is a practical approximation for submodular-style objectives. We use a coverage+novelty score:
\begin{equation}
\begin{split}
\mathrm{score}(\mathbf{T}\mid S)
&=\underbrace{|V(\mathbf{T})\setminus \cup_{\mathbf{T}'\in S}V(\mathbf{T}')|}_{\text{coverage gain}} \\
&\quad \cdot \exp\!\left(-\frac{d(\mathbf{T},\mathcal{T}_{\text{train}})}{\sigma}\right),
\end{split}
\end{equation}
where
\begin{equation}
d(\mathbf{T},\mathcal{T}_{\text{train}})=\min_{\mathbf{T}'\in\mathcal{T}_{\text{train}}}\Big(\|\mathbf{t}-\mathbf{t}'\|_2+\lambda_\psi\cdot|\mathrm{wrap}(\psi-\psi')|\Big),
\end{equation}
with camera center $\mathbf{t}$, yaw $\psi$, $\lambda_\psi=0.20$, and $\sigma=0.35$ by default. We set $\sigma$ from val-split held-out RGB frames only (no depth labels, no test frames); sensitivity diagnostics are reported in the supplement.

\paragraph{CN-Coverage greedy selection.}
Candidate poses $\Omega$ are generated from random and robot-centric perturbations around train poses. In the benchmark used here, we set a fixed $|\Omega|=1000$ per scene/seed and reuse it across methods and budgets. We first select a unique pose set size $N_{\mathrm{sel}}=\min(N,500)$ using greedy coverage+novelty (Algorithm~\ref{alg:qcoverage}). For $N>500$, training draws additional views with replacement from this selected set to reach budget $N$. Detailed sampler bounds are listed in Appendix~B.

\begin{algorithm}[H]
    \centering
    \fbox{
    \begin{minipage}{0.95\linewidth}
    \small
    \textbf{Algorithm 1: CN-Coverage (greedy)}\\
    \textbf{Input:} candidate poses $\Omega$, train poses $\mathcal{T}_{\text{train}}$, budget $N$.\\
    \textbf{Initialize:} $N_{\mathrm{sel}}\leftarrow \min(N,500)$, $S\leftarrow \varnothing$, covered voxels $\mathcal{U}\leftarrow \varnothing$.\\
    \textbf{For} $k=1..N_{\mathrm{sel}}$: for each $\mathbf{T}\in\Omega\setminus S$, compute
    \[
    \Delta(\mathbf{T}) = |V(\mathbf{T})\setminus \mathcal{U}|,\quad
    \pi(\mathbf{T}) = \exp(-d(\mathbf{T},\mathcal{T}_{\text{train}})/\sigma),
    \]
    and select $\mathbf{T}^\star=\arg\max_{\mathbf{T}} \Delta(\mathbf{T})\pi(\mathbf{T})$.\\
    Update $S\leftarrow S\cup\{\mathbf{T}^\star\}$, $\mathcal{U}\leftarrow \mathcal{U}\cup V(\mathbf{T}^\star)$.\\
    \textbf{Return:} selected set $S$ (for $N>500$, sample with replacement from $S$ to reach $N$).
    \end{minipage}
    }
    \caption{CN-Coverage greedy coverage+novelty view selection used for scaling.}
    \label{alg:qcoverage}
\end{algorithm}

\subsection{Training and safe scaling}
We train with robust inverse-depth loss:
\begin{equation}
\mathcal{L}=\sum_p V_{\mathrm{gt},p}\,\rho(\hat d_p^{-1}-d_p^{-1}) + \lambda_s\mathcal{L}_{\mathrm{smooth}} + \lambda_t\mathcal{L}_{\mathrm{temp}}.
\end{equation}

Guardrail observation mixing is secondary. We refer to this module as a \textbf{Gaussian Observation Layer (GOL)}. \textbf{GOL-Gated} samples 3DGS RGB versus mesh+hist fallback using scene-quality gate $g(q_s)$, while \textbf{GOL-Composite} blends both observations per pixel. This reduces regressions in low-quality 3DGS teachers while preserving scaling gains in high-quality scenes.

For gating, we define scene reliability from held-out 3DGS re-render quality:
\begin{equation}
q_s=\min\!\left(\frac{\mathrm{PSNR}_s}{10.0},\frac{\mathrm{SSIM}_s}{0.20},\frac{0.80}{\mathrm{LPIPS}_s}\right),
\end{equation}
and map it to GS sampling probability
\begin{equation}
g(q_s)=\operatorname{sigmoid}\!\left(k(q_s-\tau)\right), \quad k=8,\ \tau=1.
\end{equation}
During training, GOL-Gated samples GS observations with probability $g(q_s)$ and mesh+hist fallback otherwise. These quality metrics are computed from val-split held-out RGB views only and do not use depth-test labels.

\paragraph{Design choices and guarantees.}
CN-Coverage is a greedy heuristic: adding the novelty factor preserves practical stability but breaks strict submodularity guarantees. We therefore validate robustness empirically with sigma/voxel/distance sensitivity and convergence-budget checks (Appendix~C), and report full selection-cost details in Appendix~B.

\section{Experiments}
\subsection{Setup}
\paragraph{Data and protocol.}
We evaluate on 20 TUM RGBD sequences~\cite{sturm2012benchmark} from Freiburg-1 and Freiburg-3. We use fixed temporal train/val/test splits at $320\times240$ (up to 180 frames per sequence). We use Real2Render2Real supervision: 3DGS renders RGB observations, while mesh rendering provides aligned metric depth/visibility labels. The primary analysis is a step-matched scaling sweep with fixed optimization budget (\texttt{max\_train\_steps}=400). Rendered-view budgets are
\[
N\in\{0,25,50,100,200,500,1000,2000\},
\]
where $N$ is the number of additional rendered training views. We select $N_{\text{unique}}=\min(N,500)$ unique poses per scene; for $N>500$, training samples with replacement from that selected set to reach total budget $N$.
All synthetic training conditions consume rendered RGB inputs, while evaluation is always on held-out real RGB test frames with real depth for metrics.

\paragraph{Observation teacher construction and render domains.}
Our 3DGS observation teacher is built with a non-iterative splat initialization from per-scene RGB-D backprojection (voxel-downsampled points, Gaussian scales from local kNN, and per-point RGB colors), with zero gradient optimization iterations. We build this teacher from train-split frames and poses only. We use this throughput-oriented setup to run broad scaling sweeps; the trade-off is lower observation-teacher fidelity than fully optimized 3DGS fitting. We use three mesh render domains: \textbf{mesh(vertex)} (direct vertex-color rendering), \textbf{mesh+Hist} (per-scene LAB Reinhard color transfer toward train real-RGB statistics), and \textbf{mesh-shaded(sim)} (normal-based shaded simulator-style rendering).

\paragraph{Mesh oracle construction.}
For metric supervision, we build per-scene meshes with Open3D Scalable TSDF fusion from \emph{train-split only} RGB-D frames and camera poses (voxel length 0.01\,m, SDF truncation 0.04\,m, depth truncation 4.0\,m), then ray-cast depth/visibility with z-buffering. We remove tiny disconnected components after fusion to stabilize supervision and keep the metric frame aligned to dataset poses (no post-hoc scale fitting). Held-out mesh-vs-sensor checks show adequate oracle fidelity for label rendering (median absolute depth error 0.028\,m, mean 0.088\,m; full diagnostics in Appendix~A). Both the TSDF mesh oracle and the 3DGS observation teacher are constructed from train-split frames/poses only, with no test-frame usage; no test depth is used for gating.
\begin{table}[htbp]
\centering
\footnotesize
\setlength{\tabcolsep}{4pt}
\caption{Mesh-oracle fidelity.}
\label{tab:mesh_fidelity_main}
\begin{tabular}{lcc}
\toprule
Metric & Value & Split/frames \\
\midrule
Median abs depth error (m) & 0.028 $\pm$ 0.019 & held-out real frames \\
Mean abs depth error (m) & 0.088 $\pm$ 0.111 & held-out real frames \\
Valid-pixel fraction & 0.875 $\pm$ 0.171 & held-out real frames \\
\bottomrule
\end{tabular}
\end{table}

\paragraph{Model and training settings.}
We train the same depth network and losses across methods; only observation source and view-selection policy change. The student is DepthUNet (32 base channels) at $320{\times}240$, optimized with AdamW (LR $5\mathrm{e}{-5}$, batch 12, step cap 400 for scaling sweeps). Reported metric is metric AbsRel on real test RGB (lower is better), with per-sequence means and 95\% CIs.

\paragraph{Compact implementation definitions.}
Coverage $V(\mathbf{T})$ is computed by mesh-depth rendering at pose $\mathbf{T}$, z-buffer visibility masking, and voxelized backprojection of valid depth points ($\nu{=}0.10$\,m, stride $r{=}12$). Candidate pool $\Omega$ is generated once per scene/seed with random+robot perturbations and then reused across samplers. Novelty distance is
$d(\mathbf{T},\mathcal{T}_{\text{train}})=\|\Delta \mathbf{t}\|_2+\lambda_\psi |\Delta \psi|$ ($\lambda_\psi{=}0.20$, $\sigma{=}0.35$ in $\exp(-d/\sigma)$). We tune $\sigma$ on val-split RGB frames only (no depth labels, no test frames). Temporal loss uses adjacent monocular training pairs $(i,i{+}1)$ with reprojection consistency. Full model/loss settings, sampler bounds, and selection-cost details are listed in Appendix~B.
\FloatBarrier

\subsection{Main step-matched scaling result}
\begin{figure}[!htbp]
    \centering
    \includegraphics[width=0.95\linewidth]{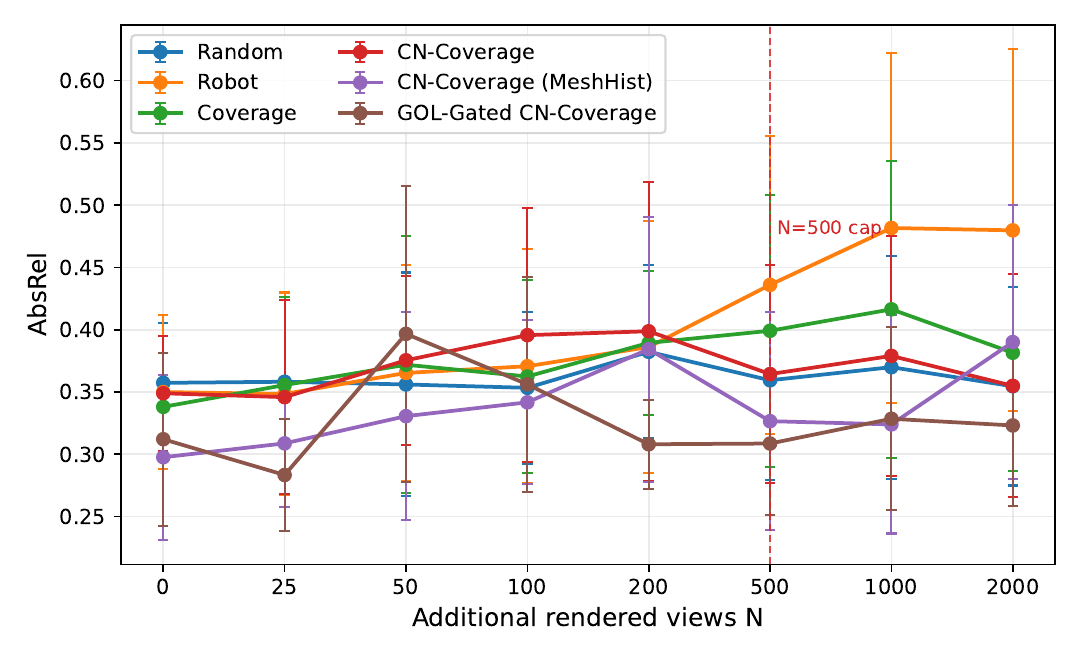}
    \caption{Step-matched scaling: metric AbsRel vs budget $N$ for Random, Robot, Coverage, CN-Coverage, CN-Coverage (MeshHist), and GOL-Gated CN-Coverage. Error bars denote 95\% CI across sequences; connecting lines are guides to the eye. Budgets are discrete and plotted on categorical x-axis positions. The dashed vertical marker at $N{=}500$ denotes the unique-view cap; for $N{>}500$, training samples with replacement from the selected set (resampling-based count scaling).}
    \label{fig:scaling_main}
\end{figure}
Figure~\ref{fig:scaling_main} and Table~\ref{tab:scaling_main} show the central result: naive view-count scaling is non-monotonic and can regress at high $N$, while structured sampling with guardrail fallback is among the strongest and most stable for medium/high budgets ($N\geq200$). Table~\ref{tab:scaling_main} reports metric AbsRel as mean $\pm$ 95\% CI across samplers and budgets. Here $N$ counts additional rendered training views on top of shared original frames, $N_{\mathrm{unique}}=\min(N,500)$ selected poses per scene, and for $N>500$ the budget is filled by sampling with replacement from that selected set. At $N=0$, Random/Robot/Coverage/CN-Coverage differences mainly reflect independent stochastic runs under the same 400-step budget, while CN-Coverage (MeshHist) and GOL-Gated CN-Coverage also differ by observation source/mixing.
\begin{table}[H]
\centering
\scriptsize
\setlength{\tabcolsep}{1.5pt}
\caption{Novel-view scaling.}
\label{tab:scaling_main}
\begin{tabularx}{\linewidth}{r*{6}{>{\centering\arraybackslash}X}}
\toprule
N & Random & Robot & Coverage & CN-Coverage & \begin{tabular}[c]{@{}c@{}}CN-Coverage\\(MeshHist)\end{tabular} & \begin{tabular}[c]{@{}c@{}}GOL-Gated\\CN-Coverage\end{tabular} \\
\midrule
0 & 0.36$\pm$0.05 & 0.35$\pm$0.06 & 0.34$\pm$0.04 & 0.35$\pm$0.05 & 0.30$\pm$0.07 & 0.31$\pm$0.07 \\
25 & 0.36$\pm$0.07 & 0.35$\pm$0.08 & 0.36$\pm$0.07 & 0.35$\pm$0.08 & 0.31$\pm$0.05 & 0.28$\pm$0.05 \\
50 & 0.36$\pm$0.09 & 0.37$\pm$0.09 & 0.37$\pm$0.10 & 0.38$\pm$0.07 & 0.33$\pm$0.08 & 0.40$\pm$0.12 \\
100 & 0.35$\pm$0.06 & 0.37$\pm$0.09 & 0.36$\pm$0.08 & 0.40$\pm$0.10 & 0.34$\pm$0.07 & 0.36$\pm$0.09 \\
200 & 0.38$\pm$0.07 & 0.39$\pm$0.10 & 0.39$\pm$0.06 & 0.40$\pm$0.12 & 0.38$\pm$0.11 & 0.31$\pm$0.04 \\
500 & 0.36$\pm$0.08 & 0.44$\pm$0.12 & 0.40$\pm$0.11 & 0.36$\pm$0.09 & 0.33$\pm$0.09 & 0.31$\pm$0.06 \\
1000 & 0.37$\pm$0.09 & 0.48$\pm$0.14 & 0.42$\pm$0.12 & 0.38$\pm$0.10 & 0.32$\pm$0.09 & 0.33$\pm$0.07 \\
2000 & 0.35$\pm$0.08 & 0.48$\pm$0.15 & 0.38$\pm$0.10 & 0.35$\pm$0.09 & 0.39$\pm$0.11 & 0.32$\pm$0.06 \\
\bottomrule
\end{tabularx}
\end{table}

We define stability as bounded medium/high-budget error (low mean, low worst-case, and low range over $N\geq200$), quantified in Table~\ref{tab:scaling_stability_main}.
\begin{table}[htbp]
\centering
\scriptsize
\setlength{\tabcolsep}{3pt}
\caption{Scaling stability summary.}
\label{tab:scaling_stability_main}
\begin{tabular}{p{0.42\linewidth}ccc}
\toprule
Method & Mean & Worst & Range \\
\midrule
GOL-Gated CN-Coverage & 0.317 & 0.328 & 0.020 \\
CN-Coverage (MeshHist) & 0.356 & 0.390 & 0.066 \\
Random & 0.367 & 0.382 & 0.028 \\
CN-Coverage & 0.374 & 0.399 & 0.044 \\
Coverage & 0.397 & 0.416 & 0.035 \\
Robot & 0.446 & 0.482 & 0.096 \\
\bottomrule
\end{tabular}
\end{table}

Table~\ref{tab:convergence_control_main} provides a fixed-$N$ convergence control (400 vs 2000 steps), reported as mean $\pm$ 95\% CI across 20 sequences with $\Delta=$AbsRel@2000$-$AbsRel@400 (negative is better): absolute errors drop with more optimization for all methods, while the qualitative stability ordering (guardrailed CN-Coverage stronger than naive/random at fixed $N$) is preserved.
\begin{table}[!b]
\centering
\scriptsize
\setlength{\tabcolsep}{2.5pt}
\caption{Convergence-budget control.}
\label{tab:convergence_control_main}
\begin{tabular}{p{0.36\linewidth}crrr}
\toprule
Method & $N$ & 400-step & 2000-step & $\Delta$ \\
\midrule
GOL-Gated CN-Coverage & 200 & 0.322 $\pm$ 0.087 & 0.253 $\pm$ 0.046 & -0.069 \\
Random & 200 & 0.369 $\pm$ 0.092 & 0.314 $\pm$ 0.046 & -0.055 \\
GOL-Gated CN-Coverage & 2000 & 0.356 $\pm$ 0.082 & 0.263 $\pm$ 0.051 & -0.093 \\
Random & 2000 & 0.362 $\pm$ 0.092 & 0.316 $\pm$ 0.042 & -0.046 \\
\bottomrule
\end{tabular}
\end{table}

Under fixed compute, we observe a practical sweet spot at small curated budgets (notably around $N=25$ for the guarded variant), before broader augmentation can induce under-training or higher-variance regimes.
At $N=0$, sampling policy is irrelevant: Random/Robot/Coverage/CN-Coverage entries are independent repeats that expose variance under identical data; CN-Coverage (MeshHist) and GOL-Gated CN-Coverage rows additionally differ by observation source/mixing. CN-Coverage selects up to 500 unique viewpoints per scene (details in Appendix~B); for $N>500$, training reuses/resamples that fixed selected pool. Beyond 500, we study resampling-based count scaling (redundancy/stochasticity), not unique-pose support expansion.
In this redundancy regime, guardrails become more important because repeated exposure can amplify low-quality observation modes if fallback mixing is absent.
We observe a variance-sensitive outlier at $N=50$ for GOL-Gated CN-Coverage; multi-seed and per-sequence diagnostics are provided in Appendix Fig.~\ref{fig:supp_n50_outlier} and the robustness tables in Appendix~C.

\subsection{Coverage lens: structure beats count}
\begin{figure}[H]
    \centering
    \includegraphics[width=0.95\linewidth]{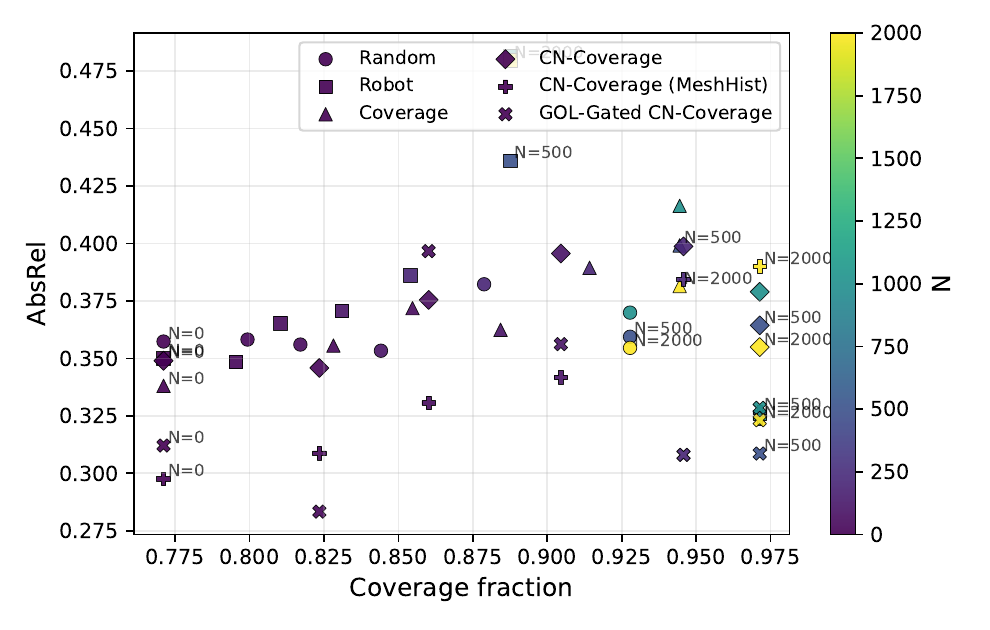}
    \caption{AbsRel versus surface-coverage fraction. Each point is one (sampler, $N$) setting.}
    \label{fig:coverage_collapse}
\end{figure}

Figure~\ref{fig:coverage_collapse} should be read as a capacity lens, not a monotonic law: coverage indicates how much surface support is exposed, but failures depend on \emph{how} that coverage is reached under novelty/extrapolation. Appendix Table~\ref{tab:scaling_coverage_corr} quantifies this: Robot/Coverage show strong positive coverage--error correlations (Pearson 0.909/0.899), while GOL-Gated CN-Coverage is near zero (Pearson 0.014), consistent with stabilized scaling. This is the key motivation for adding novelty control to pure coverage gain: coverage alone can be achieved via extrapolative poses that degrade transfer.
At $N=2000$, GOL-Gated CN-Coverage achieves the lowest absolute tail error (0.348) and avoids the large regressions observed in Coverage (0.419) and Robot (0.462). We do not claim CN-Coverage always outperforms Random in every regime; the claim is that the guarded variant maintains better high-novelty behavior.

\subsection{Tail robustness under viewpoint novelty}
\begin{figure}[!htbp]
    \centering
    \includegraphics[width=0.95\linewidth]{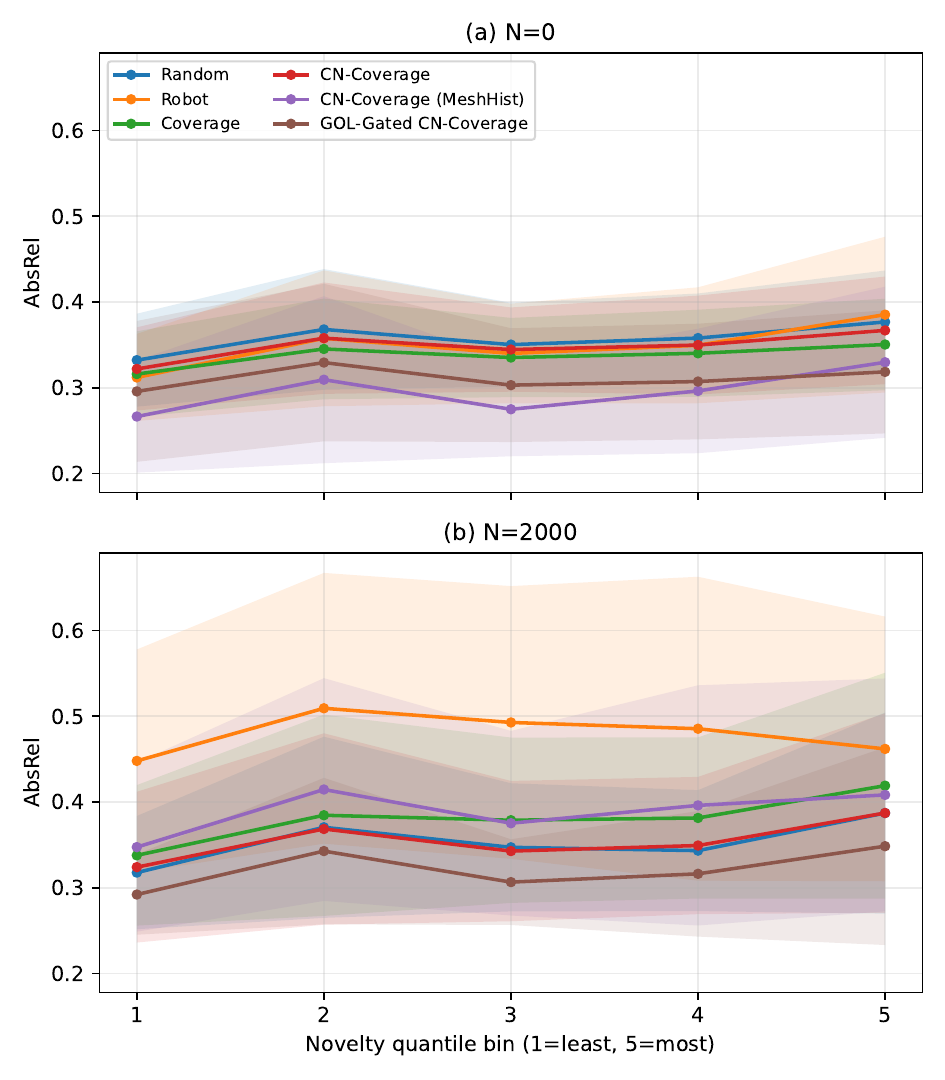}
    \caption{Error versus pose-novelty bins at low/high $N$ in vertical subpanels: (a) $N=0$, (b) $N=2000$. The x-axis uses per-sequence novelty quantile bins 1--5 (1 = least novel, 5 = most novel), so all 20 sequences contribute to each panel. Shaded regions denote 95\% CI across sequences.}
    \label{fig:tail}
\end{figure}

We evaluate robustness in the highest novelty bin because deployment failures are concentrated there. Table~\ref{tab:scaling_tail_summary} reports highest-novelty-bin (quantile bin 5/5) AbsRel at $N=0$ and $N=2000$ as mean $\pm$ 95\% CI across scenes, with $\Delta$Tail defined as Tail@$N{=}2000$ minus Tail@$N{=}0$. Figure~\ref{fig:tail} shows the full novelty-bin profiles at $N=0$ and $N=2000$ across per-sequence novelty quantile bins.
\begin{table}[!t]
\centering
\small
\setlength{\tabcolsep}{2.5pt}
\caption{Tail summary.}
\label{tab:scaling_tail_summary}
\begin{tabularx}{\linewidth}{>{\raggedright\arraybackslash}p{0.30\linewidth}>{\centering\arraybackslash}p{0.10\linewidth}*{3}{>{\centering\arraybackslash}X}}
\toprule
Mode & Scenes & Tail@0 & Tail@2000 & $\Delta$Tail \\
\midrule
CN-Coverage & 20 & 0.367 $\pm$ 0.063 & 0.387 $\pm$ 0.116 & 0.020 $\pm$ 0.082 \\
CN-Coverage (MeshHist) & 20 & 0.330 $\pm$ 0.088 & 0.408 $\pm$ 0.136 & 0.079 $\pm$ 0.062 \\
Coverage & 20 & 0.350 $\pm$ 0.053 & 0.419 $\pm$ 0.132 & 0.069 $\pm$ 0.109 \\
GOL-Gated CN-Coverage & 20 & 0.319 $\pm$ 0.072 & 0.348 $\pm$ 0.115 & 0.030 $\pm$ 0.083 \\
Random & 20 & 0.377 $\pm$ 0.060 & 0.387 $\pm$ 0.117 & 0.010 $\pm$ 0.082 \\
Robot & 20 & 0.385 $\pm$ 0.091 & 0.462 $\pm$ 0.154 & 0.077 $\pm$ 0.098 \\
\bottomrule
\end{tabularx}
\end{table}

\FloatBarrier

\subsection{Teacher quality and guardrail behavior}
\begin{figure}[!htbp]
    \centering
    \includegraphics[width=0.95\linewidth]{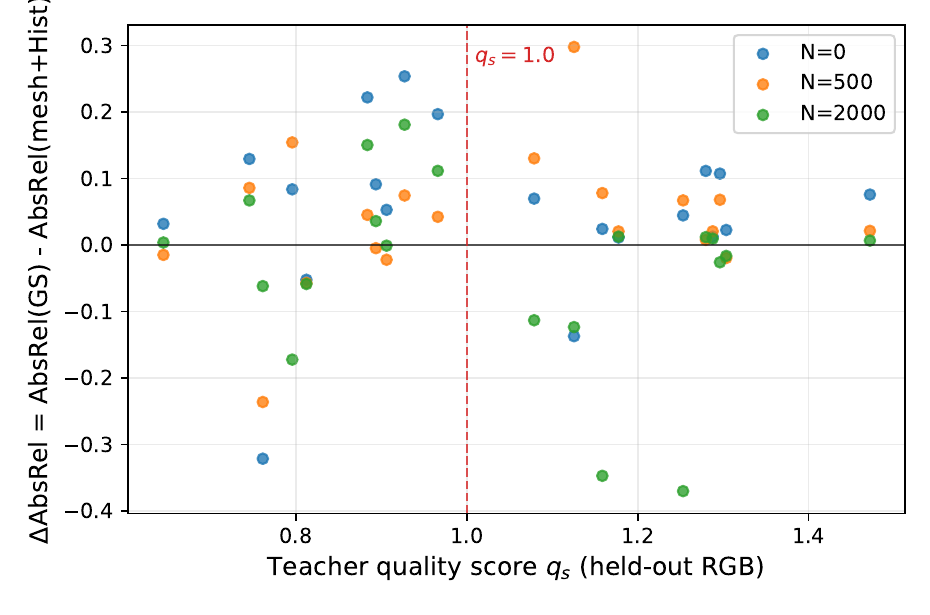}
    \caption{Teacher-quality interaction scatter at multiple budgets, where $\Delta$AbsRel $=$ AbsRel(GS) $-$ AbsRel(mesh+Hist) and negative is better. Dashed line marks gate threshold $q_s=1.0$. Correlations (Pearson/Spearman) are $N{=}0$: 0.06/$-0.07$, $N{=}500$: 0.21/0.08, $N{=}2000$: $-0.23$/$-0.11$.}
    \label{fig:quality_interact}
\end{figure}

Table~\ref{tab:teacher_protocol} summarizes the teacher-quality gating protocol and high-quality rate. Scenes are not excluded from training; the gate only controls GS-vs-fallback mixing probability. Figure~\ref{fig:quality_interact} and the appendix quality-bucket summary table show heterogeneous GS-vs-mesh behavior: at $N=2000$, high-quality scenes are more likely to benefit from GS (60\% with $\Delta$AbsRel$<0$), while low-quality scenes regress more often. This supports guardrail mixing as a risk-control layer rather than unconditional GS usage. A practical pattern emerges: observation-teacher usefulness is limited at small budgets but appears more often at large budgets, where cross-view consistency matters more. Importantly, we do not remove scenes from training; gate statistics only control per-scene GS-vs-fallback mixing from val-split held-out RGB re-render quality (no depth labels), with no test-depth leakage.
\begin{table}[H]
\centering
\small
\caption{Gating protocol.}
\label{tab:teacher_protocol}
\begin{tabular}{ll}
\toprule
Setting & Value \\
\midrule
Total scenes & 20 \\
High-quality scenes (q\_s >= 1) & 10 \\
Low-quality scenes (q\_s < 1) & 10 \\
High-quality rate & 0.500 \\
PSNR threshold & 10.000 \\
SSIM threshold & 0.200 \\
LPIPS threshold & 0.800 \\
Gate k & 8.000 \\
Gate tau & 1.000 \\
Mean gate probability & 0.541 \\
\bottomrule
\end{tabular}
\end{table}

\FloatBarrier

\subsection{Downstream Physical-AI control proxy}
\begin{figure}[!htbp]
    \centering
    \includegraphics[width=0.80\linewidth]{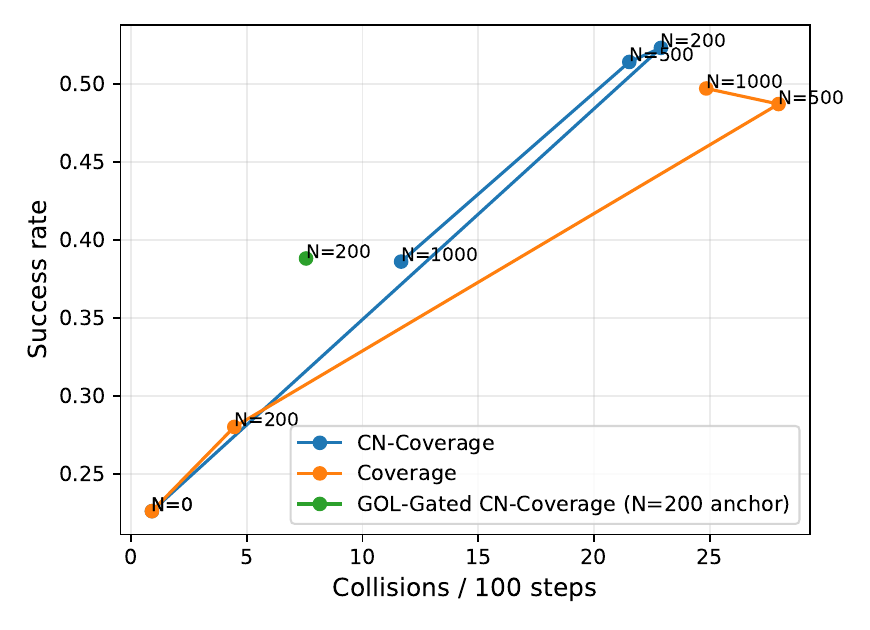}
    \caption{Success-collision Pareto across budgets $N$ (each point annotated by $N$).}
    \label{fig:downstream_pareto}
\end{figure}
\paragraph{Downstream protocol details.}
We run 1000 episodes from random start/goal pairs sampled with a minimum 1.5\,m separation. Each episode has horizon 12 actions with 0.25\,m forward steps. A motion is considered predicted-safe when predicted clearance exceeds 1.2\,m. A collision is counted when this predicted-safe motion is unsafe under sensor depth at the same threshold. Success requires zero collisions and at least 80\% oracle-safe progress toward the goal.
The downstream sweep uses depth checkpoints trained with GS observations under step-matched scaling for Coverage/CN-Coverage at $N\in\{0,200,500,1000\}$. For fair comparison, the $N=0$ point uses a shared baseline checkpoint across Coverage/CN-Coverage. We also report one GOL-Gated CN-Coverage anchor at $N=200$.
Table~\ref{tab:downstream_scaling} reports mean $\pm$ 95\% CI over 1000 episodes. Succ is zero-collision success rate, Col/100 is collisions per 100 steps, Col/fail is collisions per failed episode, and Path is progress\_moves/oracle\_safe\_moves.
\begin{table}[H]
\centering
\footnotesize
\setlength{\tabcolsep}{2pt}
\caption{Downstream control proxy.}
\label{tab:downstream_scaling}
\begin{tabularx}{\linewidth}{>{\raggedright\arraybackslash}p{0.32\linewidth}>{\centering\arraybackslash}p{0.08\linewidth}*{4}{>{\centering\arraybackslash}X}}
\toprule
Mode & N & Succ & Col/100 & Col/fail & Path \\
\midrule
CN-Coverage & 0 & 0.226 $\pm$ 0.026 & 0.900 $\pm$ 0.341 & 0.140 $\pm$ 0.053 & 0.281 $\pm$ 0.026 \\
CN-Coverage & 200 & 0.523 $\pm$ 0.031 & 22.892 $\pm$ 2.225 & 5.759 $\pm$ 0.417 & 0.647 $\pm$ 0.029 \\
CN-Coverage & 500 & 0.514 $\pm$ 0.031 & 21.525 $\pm$ 2.104 & 5.315 $\pm$ 0.394 & 0.625 $\pm$ 0.029 \\
CN-Coverage & 1000 & 0.386 $\pm$ 0.030 & 11.667 $\pm$ 1.734 & 2.280 $\pm$ 0.320 & 0.429 $\pm$ 0.029 \\
Coverage & 0 & 0.226 $\pm$ 0.026 & 0.900 $\pm$ 0.341 & 0.140 $\pm$ 0.053 & 0.281 $\pm$ 0.026 \\
Coverage & 200 & 0.280 $\pm$ 0.028 & 4.458 $\pm$ 1.037 & 0.743 $\pm$ 0.171 & 0.291 $\pm$ 0.027 \\
Coverage & 500 & 0.487 $\pm$ 0.031 & 27.975 $\pm$ 2.394 & 6.544 $\pm$ 0.397 & 0.632 $\pm$ 0.029 \\
Coverage & 1000 & 0.497 $\pm$ 0.031 & 24.850 $\pm$ 2.257 & 5.928 $\pm$ 0.396 & 0.621 $\pm$ 0.029 \\
GOL-Gated CN-Coverage* & 200 & 0.388 $\pm$ 0.030 & 7.558 $\pm$ 1.177 & 1.482 $\pm$ 0.219 & 0.507 $\pm$ 0.030 \\
\bottomrule
\end{tabularx}
\end{table}

Table~\ref{tab:downstream_scaling} and Fig.~\ref{fig:downstream_pareto} show that scaling policy, not only view count, changes success/collision trade-offs. High Col/100 can coexist with moderate success because collisions are concentrated in failed episodes, which is why Col/fail is reported separately.
This section is intentionally a stylized diagnostic, not a full RL/navigation benchmark. The objective is to test behavior-level sensitivity to depth scaling policies under controlled kinematics and map thresholds, rather than to claim policy-transfer gains. We therefore report trade-off movement (success versus collisions), not dominance claims. Across modes, moderate success can appear with high Col/100 when failures are collision-heavy and clustered in a subset of episodes.
\FloatBarrier

\subsection{Additional diagnostics}
\label{sec:diag}
Appendix~C reports exposure-matched budgets and stochastic-greedy baselines; these checks improve all methods with more compute but preserve the same qualitative ordering used in the main claims. Runtime/practicality, render-domain gap, and cross-family holdout diagnostics are reported in Appendix~A.
\FloatBarrier

\section{Discussion and Limitations}
\paragraph{Main takeaway.}
The central result is a scaling-stability result: robustness is better preserved when viewpoint support is expanded with structure (CN-Coverage plus guardrails), not when view count is increased indiscriminately. This aligns with both the pose-distribution and coverage lenses.

\paragraph{Scope and limitations.}
(1) Our setup uses simulator-style rendering + kinematic rollout, not full physics simulation. (2) We operate on mostly static indoor scenes; strong dynamics and severe pose drift still hurt teacher reliability. (3) External validity beyond TUM remains important future work. (4) Effective $N>500$ uses controlled resampling from a fixed rendered novel-view pool in this implementation; true unique-view scaling to much larger $N$ would require larger rendered pools or online rendering infrastructure.

\section{Conclusion}
We introduced \textbf{Splat2Real}, reframing 3DGS depth transfer as a \emph{novel-view scaling} problem. On a 20-sequence benchmark with step-matched sweeps, naive scaling is non-monotonic and can regress as budget grows. Relative to Robot and Coverage policies, CN-Coverage reduces worst-case scaling failures, while GOL-Gated CN-Coverage delivers the strongest medium/high-budget stability and the lowest high-novelty tail error. The results also show that larger view budgets are not sufficient by themselves: under fixed compute, small curated budgets can be strong, and scaling beyond the unique-view cap is a resampling regime where guardrails matter more.

Our analysis provides two practical insights. First, coverage is useful but not a standalone objective: high coverage reached through extrapolative poses can still degrade transfer, motivating explicit novelty control. Second, GS-vs-mesh behavior is scene-dependent; quality-aware mixing acts as risk control by limiting low-quality-teacher failures rather than assuming unconditional GS gains. Downstream control-proxy results further indicate embodied relevance by shifting success--collision trade-offs, while remaining explicitly short of policy-transfer claims. Overall, the key takeaway is that \emph{how} views are scaled dominates \emph{how many} views are added.

\FloatBarrier

\bibliographystyle{IEEEtran}
\bibliography{refs}

@article{kerbl2023gs,
  title={3D Gaussian Splatting for Real-Time Radiance Field Rendering},
  author={Kerbl, Bernhard and Kopanas, Georgios and Leimk{"u}hler, Thomas and Drettakis, George},
  journal={ACM Transactions on Graphics},
  volume={42},
  number={4},
  year={2023}
}

@article{feldmann2024nerfmentation,
  title={NeRFmentation: NeRF-based Augmentation for Monocular Depth Estimation},
  author={Feldmann, Casimir and Siegenheim, Niall and Hars, Nikolas and Rabuzin, Lovro and Ertugrul, Mert and Wolfart, Luca and Pollefeys, Marc and Bauer, Zuria and Oswald, Martin R.},
  journal={arXiv preprint arXiv:2401.03771},
  year={2024}
}

@article{xu2024depthsplat,
  title={DepthSplat: Connecting Gaussian Splatting and Depth},
  author={Xu, Haofei and Peng, Songyou and Wang, Fangjinhua and Blum, Hermann and Barath, Daniel and Geiger, Andreas and Pollefeys, Marc},
  journal={arXiv preprint arXiv:2410.13862},
  year={2024}
}

@inproceedings{qureshi2025splatsim,
  title={SplatSim: Zero-Shot Sim2Real Transfer of RGB Manipulation Policies Using Gaussian Splatting},
  author={Qureshi, Mohammad Nomaan and Garg, Sparsh and Yandun, Francisco and Held, David and Kantor, George and Silwal, Abhisesh},
  booktitle={Conference on Robot Learning (OpenReview)},
  year={2025},
  url={https://openreview.net/forum?id=UPEkp5NCx4}
}

@article{li2024robogsim,
  title={RoboGSim: A Real2Sim2Real Robotic Gaussian Splatting Simulator},
  author={Li, Xinhai and Li, Jialin and Zhang, Ziheng and Zhang, Rui and Jia, Fan and Wang, Tiancai and Fan, Haoqiang and Tseng, Kuo-Kun and Wang, Ruiping},
  journal={arXiv preprint arXiv:2411.11839},
  year={2024}
}

@article{wu2024rlgsbridge,
  title={RL-GSBridge: 3D Gaussian Splatting Based Real2Sim2Real Method for Robotic Manipulation Learning},
  author={Wu, Yuxuan and Pan, Lei and Wu, Wenhua and Wang, Guangming and Miao, Yanzi and Xu, Fan and Wang, Hesheng},
  journal={arXiv preprint arXiv:2409.20291},
  year={2024}
}

@inproceedings{chhablani2025embodiedsplat,
  title={EmbodiedSplat: Personalized Embodied Navigation in a 3D Gaussian Splatting World},
  author={Chhablani, Bhavy and Yao, Ruixi and Peri, Neehar and Dar, Amir and Khudanpur, Sanjeev and Arora, Rama and Peng, Hao and Manocha, Dinesh and Choudhury, Sanjiban and Krishna, Ranjay and Hajishirzi, Hannaneh},
  booktitle={Proceedings of the IEEE/CVF International Conference on Computer Vision},
  year={2025}
}

@inproceedings{sturm2012benchmark,
  title={A Benchmark for the Evaluation of RGB-D SLAM Systems},
  author={Sturm, J{"u}rgen and Engelhard, Nikolas and Endres, Felix and Burgard, Wolfram and Cremers, Daniel},
  booktitle={Proceedings of the IEEE/RSJ International Conference on Intelligent Robots and Systems},
  year={2012}
}

@article{ranftl2021dpt,
  title={Vision Transformers for Dense Prediction},
  author={Ranftl, Ren{\'e} and Bochkovskiy, Alexey and Koltun, Vladlen},
  journal={arXiv preprint arXiv:2103.13413},
  year={2021}
}

@article{tobin2017domain,
  title={Domain Randomization for Transferring Deep Neural Networks from Simulation to the Real World},
  author={Tobin, Josh and Fong, Rachel and Ray, Alex and Schneider, Jonas and Zaremba, Wojciech and Abbeel, Pieter},
  journal={arXiv preprint arXiv:1703.06907},
  year={2017}
}

@article{ganin2016domain,
  title={Domain-Adversarial Training of Neural Networks},
  author={Ganin, Yaroslav and Ustinova, Evgeniya and Ajakan, Hana and Germain, Pascal and Larochelle, Hugo and Laviolette, Fran{\c}ois and Marchand, Mario and Lempitsky, Victor},
  journal={Journal of Machine Learning Research},
  volume={17},
  number={59},
  pages={1--35},
  year={2016}
}

@inproceedings{su2015render,
  title={Render for CNN: Viewpoint Estimation in Images Using CNNs Trained with Rendered 3D Model Views},
  author={Su, Hao and Qi, Charles R. and Li, Yangyan and Guibas, Leonidas J.},
  booktitle={Proceedings of the IEEE International Conference on Computer Vision},
  year={2015}
}

@article{hendrycks2020augmix,
  title={AugMix: A Simple Data Processing Method to Improve Robustness and Uncertainty},
  author={Hendrycks, Dan and Mu, Norman and Cubuk, Ekin D. and Zoph, Barret and Gilmer, Justin and Lakshminarayanan, Balaji},
  journal={arXiv preprint arXiv:1912.02781},
  year={2020}
}

@article{krause2014submodular,
  title={Submodular Function Maximization},
  author={Krause, Andreas and Golovin, Daniel},
  journal={Tractability: Practical Approaches to Hard Problems},
  pages={71--104},
  year={2014}
}

@inproceedings{bircher2016receding,
  title={Receding Horizon ``Next-Best-View'' Planner for 3D Exploration},
  author={Bircher, Andreas and Kamel, Mina and Alexis, Kostas and Oleynikova, Helen and Siegwart, Roland},
  booktitle={Proceedings of the IEEE International Conference on Robotics and Automation},
  year={2016}
}

@article{hussein2017imitation,
  title={Imitation Learning: A Survey of Learning Methods},
  author={Hussein, Ahmed and Gaber, Mohamed Medhat and Elyan, Eyad and Jayne, Chris},
  journal={ACM Computing Surveys},
  volume={50},
  number={2},
  pages={21:1--21:35},
  year={2017}
}

@article{osa2018algorithmic,
  title={An Algorithmic Perspective on Imitation Learning},
  author={Osa, Takayuki and Pajarinen, Joni and Neumann, Gerhard and Bagnell, J. Andrew and Abbeel, Pieter and Peters, Jan},
  journal={Foundations and Trends in Robotics},
  volume={7},
  number={1--2},
  pages={1--179},
  year={2018}
}

@article{hinton2015distilling,
  title={Distilling the Knowledge in a Neural Network},
  author={Hinton, Geoffrey and Vinyals, Oriol and Dean, Jeff},
  journal={arXiv preprint arXiv:1503.02531},
  year={2015}
}

\clearpage
\setcounter{section}{0}
\setcounter{subsection}{0}
\setcounter{figure}{0}
\setcounter{table}{0}
\renewcommand{\thesection}{Appendix \Alph{section}}
\renewcommand{\thesubsection}{\thesection.\arabic{subsection}}
\renewcommand{\thefigure}{\Alph{section}.\arabic{figure}}
\renewcommand{\thetable}{\Alph{section}.\arabic{table}}
\renewcommand{\theHsection}{appendix.\thesection}
\renewcommand{\theHfigure}{appendix.\Alph{section}.\arabic{figure}}
\renewcommand{\theHtable}{appendix.\Alph{section}.\arabic{table}}
\newcommand{\appendixsection}[1]{%
    \clearpage
    \refstepcounter{section}%
    \setcounter{figure}{0}%
    \setcounter{table}{0}%
    \twocolumn[{%
        \centering
        \vspace*{-1.2\baselineskip}%
        {\normalfont\large\scshape \thesection.\ #1\par}%
        \vspace{2.4em}%
    }]%
}

\appendixsection{Additional Statistical Tables}
This appendix complements the main paper with expanded statistics, reproducibility details, robustness diagnostics, and per-sequence analyses that would interrupt the main scaling narrative. Appendix~A starts with the statistical tables that most directly support the headline claims; later appendices provide implementation details, stress tests, additional figures, and paired summaries derived from the same frozen result logs.

\begin{figure}[!htbp]
    \centering
    \includegraphics[width=0.86\linewidth]{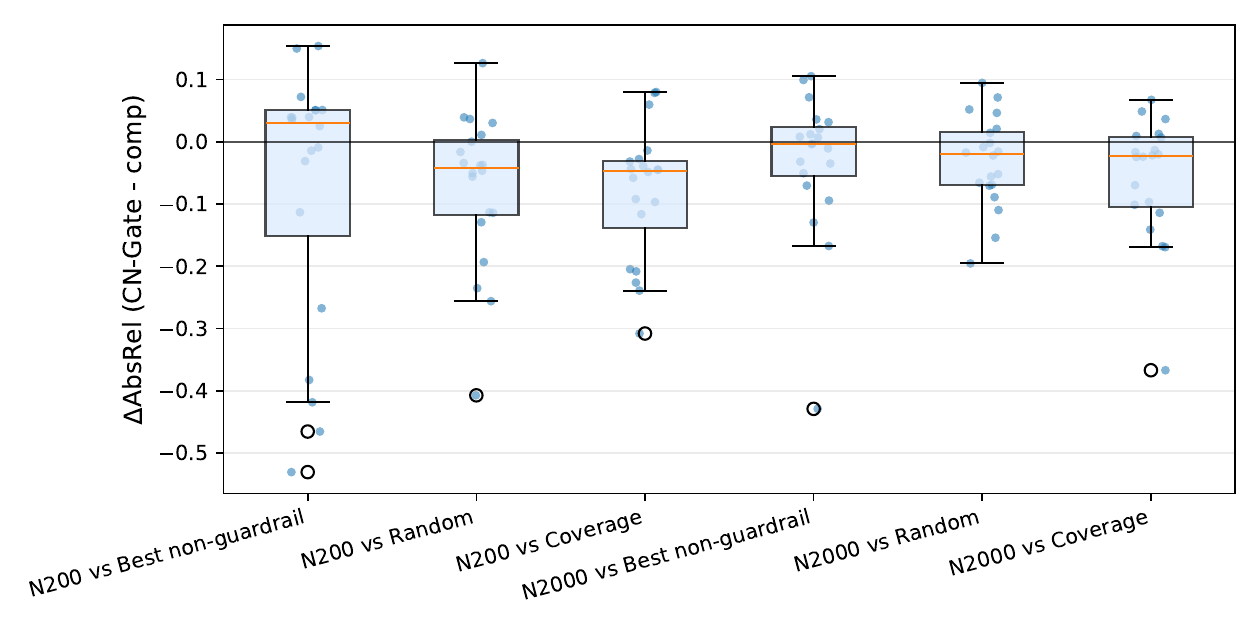}
    \caption{Per-sequence paired $\Delta$AbsRel distributions at $N=\{200,2000\}$.}
    \label{fig:supp_scaling_delta}
\end{figure}

\paragraph{Claim-to-evidence map.}
\begin{itemize}
\item \textbf{Guarded scaling is the primary supported claim.} Main Fig.~2 and Tables~2--4 show the step-matched sweep, stability summary, and compute control; Appendix Tables~\ref{tab:scaling_significance}, \ref{tab:exposure_control}, \ref{tab:stochastic_baseline}, and Fig.~\ref{fig:supp_scaling_delta} add paired tests, exposure-matched controls, and a stochastic-greedy baseline.
\item \textbf{High-novelty robustness is supported more strongly than mean-error gains.} Main Fig.~4 and Table~5 show the novelty-tail behavior; Appendix Table~\ref{tab:tail_pairwise} adds paired highest-novelty comparisons against CN-Coverage and CN-Coverage (MeshHist).
\item \textbf{Coverage is diagnostic, not sufficient.} Main Fig.~3 visualizes coverage as an organizing axis; Appendix Table~\ref{tab:scaling_coverage_corr} quantifies that Robot/Coverage policies correlate strongly with worsening error, while the guarded variant does not.
\item \textbf{$q_s$ is a coarse risk-control indicator.} Main Fig.~5 and Table~6 define the gate and its threshold; Appendix Table~\ref{tab:quality_bucket_summary} and Fig.~\ref{fig:qs_threshold} show the threshold split, gate mapping, and heterogeneous scene-level behavior without claiming a strong linear predictor.
\item \textbf{The downstream section is a stylized control proxy.} Main Table~7 and Fig.~6 report trade-off movement; Appendix Fig.~\ref{fig:supp_downstream_scaling} provides the full curves and the shared-$N{=}0$ note, but does not establish policy transfer or real-robot performance.
\end{itemize}

This section collects the full statistical tables that underpin the main-paper claims. It provides reference baselines, significance tests, and additional diagnostics in one place.

\paragraph{Reference schedule baseline.}
The main paper focuses on step-matched scaling. Table~\ref{tab:main_sim2real_metric} reports the non-step-matched reference protocol for context, where synthetic render-domain baselines are compared under longer training. It uses metric depth without median scaling, reports mean $\pm$ std across scenes, and highlights the best/second-best synthetic render-domain conditions while excluding the real upper bound from ranking.
\begin{table}[H]
\centering
\scriptsize
\setlength{\tabcolsep}{1.7pt}
\caption{Sim2Real results.}
\label{tab:main_sim2real_metric}
\resizebox{\columnwidth}{!}{%
\begin{tabular}{lccccc}
\toprule
Condition & AbsRel & RMSE & SIlog & Scale & d1 \\
\midrule
real (upper) & 0.221 $\pm$ 0.086 & 0.505 $\pm$ 0.171 & 21.556 & 0.163 & 0.605 $\pm$ 0.160 \\
mesh & 0.283 $\pm$ 0.104 & 0.583 $\pm$ 0.213 & 24.000 & 0.226 & 0.506 $\pm$ 0.160 \\
mesh+DR & 0.308 $\pm$ 0.098 & 0.621 $\pm$ 0.213 & 25.575 & 0.255 & 0.458 $\pm$ 0.180 \\
mesh+Hist & 0.281 $\pm$ 0.149 & \textbf{0.527 $\pm$ 0.150} & \underline{22.192} & 0.221 & \textbf{0.560 $\pm$ 0.187} \\
mesh+AdaIN & \underline{0.279 $\pm$ 0.115} & 0.563 $\pm$ 0.155 & 24.724 & \underline{0.220} & \underline{0.523 $\pm$ 0.145} \\
3DGS & 0.315 $\pm$ 0.073 & 0.726 $\pm$ 0.239 & 27.978 & 0.257 & 0.379 $\pm$ 0.168 \\
GOL-Gated & \textbf{0.275 $\pm$ 0.123} & 0.575 $\pm$ 0.274 & \textbf{21.424} & \textbf{0.217} & 0.502 $\pm$ 0.224 \\
GOL-Composite & 0.320 $\pm$ 0.168 & \underline{0.562 $\pm$ 0.132} & 22.654 & 0.275 & 0.495 $\pm$ 0.216 \\
\bottomrule
\end{tabular}
}
\end{table}

This table establishes that, under the longer reference schedule, the strongest synthetic condition is guarded rather than raw 3DGS, and that appearance baselines such as mesh+Hist and mesh+AdaIN remain competitive. It does not establish the step-matched scaling claim, because the schedule and observation budget differ from the main-paper sweep.

\paragraph{Paired significance tests.}
Table~\ref{tab:scaling_significance} reports scene-level Wilcoxon tests for GOL-Gated CN-Coverage against key comparators at $N=\{200,2000\}$. $\Delta$AbsRel is target minus comparator, so negative is better. Here ``Best non-guardrail'' means a per-sequence oracle baseline: the minimum AbsRel among \{Random, Robot, Coverage, CN-Coverage, CN-Coverage (MeshHist)\} at the same $N$.
\begin{table}[H]
\centering
\footnotesize
\caption{Paired Wilcoxon tests.}
\label{tab:scaling_significance}
\begin{tabular}{rlrlll}
\toprule
N & Comp & Scenes & MedD & MeanD & p \\
\midrule
200 & Best non-guardrail & 20 & 0.030 & -0.076 & 0.869 \\
200 & Coverage & 20 & -0.047 & -0.081 & 0.00558 \\
200 & Random & 20 & -0.042 & -0.074 & 0.00944 \\
2000 & Best non-guardrail & 20 & -0.003 & -0.032 & 0.498 \\
2000 & Coverage & 20 & -0.023 & -0.058 & 0.0121 \\
2000 & Random & 20 & -0.020 & -0.031 & 0.0826 \\
\bottomrule
\end{tabular}
\end{table}

These tests establish that the guarded variant is significantly better than Coverage at both budgets and better than Random at $N=200$. They do not establish a universal per-sequence win over every comparator; direct guardrail-vs-CN and guardrail-vs-MeshHist paired summaries are added later in Table~\ref{tab:guardrail_pairwise}.

\paragraph{Coverage-oriented summaries.}
Table~\ref{tab:scaling_coverage_corr} quantifies how coverage fraction correlates with error across samplers. Table~\ref{tab:scaling_summary} reports the best sampler per budget to complement the full scaling table in the main paper.
\begin{table}[H]
\centering
\small
\caption{Coverage correlation.}
\label{tab:scaling_coverage_corr}
\begin{tabular}{lrrr}
\toprule
Mode & NPoints & Pearson & Spearman \\
\midrule
All & 48 & 0.182 & 0.182 \\
Random & 8 & 0.296 & 0.244 \\
Robot & 8 & 0.909 & 0.952 \\
Coverage & 8 & 0.899 & 0.903 \\
CN-Coverage & 8 & 0.452 & 0.390 \\
CN-Coverage (MeshHist) & 8 & 0.658 & 0.537 \\
GOL-Gated CN-Coverage & 8 & 0.014 & 0.098 \\
\bottomrule
\end{tabular}
\end{table}

\begin{table}[H]
\centering
\scriptsize
\setlength{\tabcolsep}{3pt}
\caption{Scaling summary.}
\label{tab:scaling_summary}
\resizebox{\columnwidth}{!}{%
\begin{tabular}{rlll}
\toprule
N & BestSampler & BestAbsRel & Coverage \\
\midrule
0 & CN-Coverage (MeshHist) & 0.298 $\pm$ 0.066 & 0.771 \\
25 & GOL-Gated CN-Coverage & 0.283 $\pm$ 0.045 & 0.823 \\
50 & CN-Coverage (MeshHist) & 0.331 $\pm$ 0.084 & 0.860 \\
100 & CN-Coverage (MeshHist) & 0.342 $\pm$ 0.066 & 0.905 \\
200 & GOL-Gated CN-Coverage & 0.308 $\pm$ 0.036 & 0.946 \\
500 & GOL-Gated CN-Coverage & 0.309 $\pm$ 0.058 & 0.971 \\
1000 & CN-Coverage (MeshHist) & 0.324 $\pm$ 0.088 & 0.971 \\
2000 & GOL-Gated CN-Coverage & 0.323 $\pm$ 0.065 & 0.971 \\
\bottomrule
\end{tabular}
}
\end{table}

These summaries establish that coverage is a useful diagnostic axis and that the guarded variant is competitive at multiple budgets. They do not imply that maximizing coverage alone is sufficient: positive coverage--error correlations for Robot/Coverage indicate that extrapolative views can raise both coverage and error.

\paragraph{Cross-family holdout sanity check.}
Table~\ref{tab:external_validity} reports the freiburg1$\rightarrow$freiburg3 holdout setting referenced in the main paper under the step-matched protocol at $N=200$, with all methods using the same Coverage pose policy.
\begin{table}[H]
\centering
\small
\caption{Holdout check.}
\label{tab:external_validity}
\begin{tabular}{lrrr}
\toprule
Method & AbsRel & RMSE & d1 \\
\midrule
mesh+Hist & 0.302 & 0.759 & 0.354 \\
3DGS & 0.452 & 1.107 & 0.152 \\
GOL-Gated & 0.363 & 0.882 & 0.263 \\
\bottomrule
\end{tabular}
\end{table}

This table establishes that the guardrail helps relative to raw 3DGS under cross-family shift. It does not establish external superiority over the strongest mesh baseline, so the external-validity claim remains a narrow sanity check.

\paragraph{Teacher-quality gating protocol.}
The gate thresholds and scene-quality bucket counts used for GOL-Gated mixing are reported in main Table~6. Scenes are \emph{not} excluded from training; quality buckets only modulate GS-vs-fallback sampling probability. Table~\ref{tab:quality_bucket_summary} reports the derived bucket summary at $N=\{0,500,2000\}$, where $\Delta$AbsRel $=$ AbsRel(3DGS)$-$AbsRel(mesh+Hist) and lower is better; $\pm$ denotes std across scenes.
\begin{table}[H]
\centering
\scriptsize
\setlength{\tabcolsep}{2pt}
\caption{Teacher-quality buckets.}
\label{tab:quality_bucket_summary}
\resizebox{\columnwidth}{!}{%
\begin{tabular}{rllrlll}
\toprule
N & Bucket & Qual. & Scenes & Mean$\pm$std & Median & Frac($\Delta<0$) \\
\midrule
0 & High-q & $q_s\geq1$ & 10 & 0.034 $\pm$ 0.071 & 0.034 & 0.10 \\
0 & Low-q & $q_s<1$ & 10 & 0.069 $\pm$ 0.166 & 0.087 & 0.20 \\
500 & High-q & $q_s\geq1$ & 10 & 0.069 $\pm$ 0.091 & 0.044 & 0.10 \\
500 & Low-q & $q_s<1$ & 10 & 0.007 $\pm$ 0.105 & 0.019 & 0.50 \\
2000 & High-q & $q_s\geq1$ & 10 & -0.096 $\pm$ 0.147 & -0.021 & 0.60 \\
2000 & Low-q & $q_s<1$ & 10 & 0.025 $\pm$ 0.107 & 0.020 & 0.40 \\
\bottomrule
\end{tabular}
}
\end{table}

These tables establish the exact held-out val-RGB-only thresholding protocol and show that high-quality scenes are more likely to help at large budgets. They do not establish $q_s$ as a strong linear predictor; throughout this appendix we interpret $q_s$ as a coarse risk-control indicator.

\paragraph{Render-domain and runtime diagnostics.}
Table~\ref{tab:domain_gap} reports the render-domain appearance gap against matched real frames, clarifying why guardrail fallback remains necessary in low-quality scenes. Table~\ref{tab:runtime} reports practical throughput and timing for teacher build/rendering and student training/inference; teacher-build timing uses $n{=}19$ because one sequence lacked complete timestamp logs for wall-clock reconstruction, while all reported accuracy experiments still use 20 sequences.
\begin{table}[H]
\centering
\scriptsize
\setlength{\tabcolsep}{2pt}
\caption{Domain gap.}
\label{tab:domain_gap}
\begin{tabular}{lrlll}
\toprule
Domain & Frames & LPIPS & PSNR & SSIM \\
\midrule
3DGS & 720 & 0.677 $\pm$ 0.075 & 10.855 $\pm$ 2.690 & 0.373 $\pm$ 0.151 \\
mesh(vertex) & 720 & 0.522 $\pm$ 0.105 & 11.374 $\pm$ 3.393 & 0.442 $\pm$ 0.153 \\
mesh+Hist & 720 & 0.499 $\pm$ 0.088 & 13.524 $\pm$ 3.166 & 0.506 $\pm$ 0.187 \\
mesh-shaded(sim) & 720 & 0.767 $\pm$ 0.053 & 5.863 $\pm$ 1.342 & 0.341 $\pm$ 0.161 \\
\bottomrule
\end{tabular}
\end{table}

\begin{table}[H]
\centering
\small
\setlength{\tabcolsep}{4pt}
\caption{Runtime.}
\label{tab:runtime}
\begin{tabularx}{\linewidth}{>{\raggedright\arraybackslash}p{0.36\linewidth}>{\raggedright\arraybackslash}X>{\centering\arraybackslash}p{0.16\linewidth}}
\toprule
Component & Value & Units \\
\midrule
GPU & NVIDIA GeForce RTX 5090 & -- \\
Teacher build time (n=19) & 2.529 $\pm$ 0.392 & s/sequence \\
3DGS setup & non-iterative init; voxel $=0.020$ m, points $=45000$, iters $=0$ & -- \\
GS render throughput & 219.89 & views/s \\
Student training time (reference, 6 epochs) & 36.22 & s \\
Student inference speed (320x240) & 552.56 & FPS \\
\bottomrule
\end{tabularx}
\end{table}

These diagnostics establish that 3DGS is not always the closest single-view render domain and that the reported sweeps are practically runnable with a non-iterative observation teacher. They do not prove that per-image domain metrics explain scaling behavior, nor that the runtime numbers would transfer to a fully optimized 3DGS pipeline.

\paragraph{Mesh-oracle fidelity diagnostics.}
Table~\ref{tab:mesh_fidelity} reports held-out real test-pose agreement between mesh-rendered depth and sensor depth as per-scene mean $\pm$ std across 20 sequences.
\begin{table}[H]
\centering
\scriptsize
\setlength{\tabcolsep}{3pt}
\caption{Mesh fidelity.}
\label{tab:mesh_fidelity}
\begin{tabular}{rlll}
\toprule
Scenes & MedianErr (m) & MeanErr (m) & ValidFrac \\
\midrule
20 & 0.028 $\pm$ 0.019 & 0.088 $\pm$ 0.111 & 0.875 $\pm$ 0.171 \\
\bottomrule
\end{tabular}
\end{table}

This table establishes that the TSDF mesh is accurate enough to serve as a metric supervision oracle on held-out frames. It does not imply oracle perfection; residual depth mismatch remains part of the supervision noise budget.

\appendixsection{Reproducibility Details}
This section provides the exact implementation settings needed to reproduce training and view selection. It complements the compact protocol summary in the main paper with full parameter tables.
\paragraph{Notation glossary.}
\begin{center}
\footnotesize
\begin{tabularx}{0.98\linewidth}{>{\raggedright\arraybackslash}p{0.22\linewidth}X}
\toprule
Symbol / term & Meaning \\
\midrule
$\text{3DGS / GS}$ & The 3D Gaussian Splatting observation renderer used for synthetic RGB inputs; ``GS'' is used as a compact abbreviation in some tables and run names. \\
$V_{\mathrm{gt}}(\mathbf{T})$ & Visibility mask used for supervision at pose $\mathbf{T}$, obtained by ray-casting the mesh oracle. \\
$V(\mathbf{T})$ & Coverage set used by CN-Coverage: the visible voxel set obtained from mesh-depth rendering, z-buffering, and voxelized backprojection. \\
$\Omega$ & Fixed candidate pose pool for one scene/seed. In this benchmark, $|\Omega|=1000$ and the same pool is reused across samplers and budgets. \\
$N,\;N_{\mathrm{unique}}$ & Additional rendered training views and the unique selected-pose cap. We use $N_{\mathrm{unique}}=\min(N,500)$ and resample with replacement for $N>500$. \\
$d(\mathbf{T},\mathcal{T}_{\mathrm{train}})$ & Pose distance used in Eq.~(4): $\|\Delta\mathbf{t}\|_2+\lambda_{\psi}|\Delta\psi|$, with $\Delta\psi$ in radians and $\lambda_{\psi}=0.20$ m/rad. \\
$\sigma$ & Novelty bandwidth in $\exp(-d/\sigma)$. The available results include a val-split held-out RGB-only sweep around the benchmark default $\sigma=0.35$, but not a separate automated selection log. \\
$q_s$ & Scene-level teacher quality score from held-out val-split RGB-only re-renders: $\min(\mathrm{PSNR}/10,\mathrm{SSIM}/0.20,0.80/\mathrm{LPIPS})$. \\
ScaleAbsRel & Absolute median scale-ratio error, $|\mathrm{median}(\hat d)/\mathrm{median}(d)-1|$; lower indicates better metric-scale calibration. \\
Path & Downstream path-ratio proxy summarizing safe progress toward the goal; the exact ratio definition is given in main Table~7. \\
NovQ80 & 80th percentile of frame-level novelty distance within a sequence/run, used only in the $N{=}50$ outlier table. \\
\bottomrule
\end{tabularx}
\end{center}

\begin{table}[!htbp]
\centering
\scriptsize
\setlength{\tabcolsep}{2pt}
\caption{Method matrix.}
\label{tab:method_matrix}
\begin{tabularx}{\linewidth}{>{\raggedright\arraybackslash}p{0.18\linewidth}>{\raggedright\arraybackslash}p{0.17\linewidth}>{\raggedright\arraybackslash}p{0.26\linewidth}>{\raggedright\arraybackslash}X}
\toprule
Name & Sampler & Observation source & Mixing rule \\
\midrule
Random / Robot / Coverage & corresponding pose policy & 3DGS RGB observations & none \\
CN-Coverage & coverage + bounded-novelty greedy policy & 3DGS RGB observations & none \\
CN-Coverage (MeshHist) & CN-Coverage & mesh RGB with per-scene LAB Reinhard transfer to train real-RGB statistics & none \\
mesh+AdaIN & Coverage (reference schedule only) & mesh RGB with AdaIN style transfer using a sampled train-split real frame as style reference & none \\
GOL-Gated CN-Coverage & CN-Coverage & 3DGS or mesh+Hist RGB & Bernoulli scene-level mixing with probability $g(q_s)$ \\
GOL-Composite & Coverage (reference schedule only) & 3DGS and mesh+Hist RGB & per-pixel confidence blend $c\cdot$3DGS$+(1-c)\cdot$mesh+Hist \\
\bottomrule
\end{tabularx}
\end{table}

\paragraph{Protocol clarifications carried by the main paper.}
Two visibility objects appear in the submission and are intentionally distinct: $V_{\mathrm{gt}}(\mathbf{T})$ is the oracle supervision mask from mesh ray-casting, while $V(\mathbf{T})$ is the voxelized coverage set used only for CN-Coverage scoring. Likewise, pose-sampler tables report human-readable jitter ranges in degrees, but the novelty distance converts yaw to radians internally, so $\lambda_\psi=0.20$ should be read as meters per radian. For $N=0$, Random/Robot/Coverage/CN-Coverage in the main scaling table are independent stochastic repeats under identical data; for $N>500$, the benchmark studies reuse/resampling from a fixed 500-pose selected set rather than continued support expansion.

\paragraph{Training, losses, and view-sampling protocol.}
Tables~\ref{tab:impl_details}--\ref{tab:coverage_params_and_cost} provide the compact implementation settings referenced from the main setup section.
\begin{table}[H]
\centering
\scriptsize
\setlength{\tabcolsep}{3pt}
\caption{Training details.}
\label{tab:impl_details}
\begin{tabular}{p{0.28\linewidth}p{0.66\linewidth}}
\toprule
Setting & Value \\
\midrule
Student model & DepthUNet (ch=32), single depth head \\
Input resolution & 320x240 RGB \\
Optimizer & AdamW \\
Batch size & 12 \\
Learning rate & 5.0e-05 \\
Weight decay & 1.0e-04 \\
Ref schedule & 4 epochs (no step cap) \\
Scaling schedule & step-matched, max steps=400 \\
Grad clip & global norm 1.0 \\
Depth loss & vis-masked inv-depth Charbonnier \\
Smoothness & edge-aware, $\lambda_s$=0.05 \\
Temporal loss & reprojection consistency, $\lambda_t$=0.15 \\
Temporal pairs & adjacent train frames (i,i+1) \\
DR baseline & gamma/gain/WB/blur/noise/JPEG, strength=0.10 \\
\bottomrule
\end{tabular}
\end{table}

\begin{table}[H]
\centering
\scriptsize
\setlength{\tabcolsep}{3pt}
\caption{Pose-sampler protocol.}
\label{tab:pose_sampler_params}
\begin{tabular}{p{0.24\linewidth}p{0.70\linewidth}}
\toprule
Sampler & Pose rule \\
\midrule
Random & train anchor + trans jitter N(0,[0.12,0.12,0.05] m) + yaw jitter N(0,10 deg) \\
Robot & arc offsets: radius U[0.15,0.35] m, heading U[-45,45] deg, z jitter +/-0.05 m \\
Coverage & greedy max new covered voxels \\
CN-Coverage & greedy max coverage-gain * exp(-d/sigma) \\
Candidate pool $\Omega$ & benchmark default: random+robot candidates, $|\Omega|=1000$ per scene \\
Reuse & $\Omega$ fixed per scene/seed; shared across methods and all budgets $N$ \\
Distance $d(\mathbf{T},\mathcal{T}_{\mathrm{train}})$ & $\|\Delta \mathbf{t}\|_2 + \lambda_\psi |\Delta \psi|$, with $\lambda_\psi=0.20$ \\
Yaw units & pose-sampling jitter is shown in degrees for readability, but $\Delta\psi$ is converted to radians inside $d(\cdot)$ \\
\bottomrule
\end{tabular}
\end{table}

\begin{table}[H]
\centering
\scriptsize
\setlength{\tabcolsep}{3pt}
\caption{Coverage and selection cost.}
\label{tab:coverage_params_and_cost}
\begin{tabular}{p{0.28\linewidth}p{0.66\linewidth}}
\toprule
Setting & Value \\
\midrule
Coverage set V(T) & visible voxels from mesh-depth backprojection (z-buffer visibility) \\
Visibility label $V_{\mathrm{gt}}(T)$ & supervision mask from mesh ray-casting; distinct from coverage set $V(T)$ \\
Voxel size & 0.100 m \\
Depth stride & 12 \\
$|\Omega|$ / scene & 1000 $\pm$ 0 \\
Unique selected/scene & 500 \\
Resampling regime & $N_{\mathrm{unique}}=\min(N,500)$; for $N>500$, training samples with replacement from the selected set \\
Voxel extract time & 0.248 +/- 0.038 s \\
Greedy select time & 4.316 +/- 3.513 s \\
Total select time & 4.933 +/- 3.549 s \\
Per-candidate score & 4.93 ms \\
Scene norm & coverage fraction normalized by per-scene union over all samplers \\
\bottomrule
\end{tabular}
\end{table}

These tables establish the concrete training defaults, sampler bounds, and selection-cost constants used in the submission. They do not, by themselves, validate alternative distance definitions or a unique optimum for $\sigma$; those questions are treated conservatively in Appendix~C.

\paragraph{TSDF mesh construction parameters.}
The oracle mesh is built with Open3D Scalable TSDF fusion on train-split RGB-D frames and poses (voxel length 0.01\,m, SDF truncation 0.04\,m, depth truncation 4.0\,m, RGB8 color integration), then exported as a triangle mesh and ray-cast for depth/visibility supervision. The corresponding mesh-fidelity summary is reported in the main paper setup section. This subsection establishes the oracle-construction pipeline and the train/val/test isolation used by both the mesh oracle and the 3DGS observation teacher. It does not establish that any test frames or depth labels are used for sigma or gate selection; those remain val-RGB-only diagnostics.

\appendixsection{Robustness Diagnostics}
This section tests whether the reported scaling conclusions are sensitive to seeds, hyperparameters, or budget-control choices. The goal is to separate stable trends from localized failure regimes.

\paragraph{N=50 instability regime.}
The main text notes a noisy regime around $N=50$. Table~\ref{tab:n50_seed_stability} and Table~\ref{tab:n50_outliers} isolate this behavior with seed-level variability and the worst affected sequences. In Table~\ref{tab:n50_outliers}, $\Delta$AbsRel $=$ AbsRel(Gated)$-$AbsRel(Coverage), so positive values indicate guarded regressions relative to Coverage.
\begin{table}[H]
\centering
\small
\caption{Seed stability.}
\label{tab:n50_seed_stability}
\begin{tabular}{ll}
\toprule
Seed & AbsRel \\
\midrule
7 & 0.397 $\pm$ 0.119 \\
11 & 0.339 $\pm$ 0.086 \\
19 & 0.328 $\pm$ 0.079 \\
Across-seed mean ± std & 0.354 $\pm$ 0.037 \\
\bottomrule
\end{tabular}
\end{table}

\begin{table}[H]
\centering
\scriptsize
\setlength{\tabcolsep}{2pt}
\caption{N=50 regressions.}
\label{tab:n50_outliers}
\begin{tabularx}{\linewidth}{>{\raggedright\arraybackslash}p{0.34\linewidth}>{\centering\arraybackslash}p{0.14\linewidth}>{\centering\arraybackslash}p{0.12\linewidth}>{\centering\arraybackslash}p{0.16\linewidth}>{\centering\arraybackslash}p{0.16\linewidth}}
\toprule
Scene & $\Delta$AbsRel & $q_s$ & Cov@50 & NovQ80 \\
\midrule
f1-xyz & 0.220 & 1.253 & 0.872 & 0.218 \\
f3-struct-notex-near & 0.187 & 0.761 & 0.931 & 1.368 \\
f1-desk & 0.119 & 1.158 & 0.862 & 0.336 \\
\bottomrule
\end{tabularx}
\end{table}

These tables establish that the $N=50$ spike is variance-sensitive and concentrated in a small subset of sequences. They do not establish a universal failure of the guarded method, and NovQ80 should be read only as the sequence-level 80th percentile of frame novelty distance in this diagnosis.

\paragraph{Hyperparameter and budget stress tests.}
The following tables test whether conclusions depend on one particular setting. Table~\ref{tab:voxel_sensitivity} varies coverage voxel size around the benchmark default $\nu=0.10$\,m from Table~\ref{tab:coverage_params_and_cost}; it should be read as a sanity sweep, not a replacement default. Table~\ref{tab:sigma_sensitivity} varies novelty bandwidth around the benchmark default $\sigma=0.35$ using val-split held-out RGB-only diagnostics, and Table~\ref{tab:geometry_proxy_corr} reports an auxiliary geometry-aware quality proxy correlation.
\begin{table}[H]
\centering
\scriptsize
\setlength{\tabcolsep}{2.5pt}
\caption{Voxel sensitivity.}
\label{tab:voxel_sensitivity}
\begin{tabular}{p{0.42\linewidth}crrc}
\toprule
Method & $N$ & voxel (m) & AbsRel & scenes \\
\midrule
GOL-Gated CN-Coverage & 200 & 0.010 & 0.310 $\pm$ 0.071 & 20 \\
GOL-Gated CN-Coverage & 200 & 0.020 & 0.307 $\pm$ 0.067 & 20 \\
GOL-Gated CN-Coverage & 200 & 0.050 & 0.308 $\pm$ 0.068 & 20 \\
\bottomrule
\end{tabular}
\end{table}

\begin{table}[H]
\centering
\scriptsize
\setlength{\tabcolsep}{2pt}
\caption{$\sigma$ sensitivity.}
\label{tab:sigma_sensitivity}
\begin{tabular}{p{0.42\linewidth}crrc}
\toprule
Method & $N$ & $\sigma$ & AbsRel & scenes \\
\midrule
GOL-Gated CN-Coverage & 200 & 0.170 & 0.305 $\pm$ 0.063 & 20 \\
GOL-Gated CN-Coverage & 200 & 0.350 & 0.313 $\pm$ 0.038 & 20 \\
GOL-Gated CN-Coverage & 200 & 0.700 & 0.315 $\pm$ 0.088 & 20 \\
GOL-Gated CN-Coverage & 2000 & 0.170 & 0.332 $\pm$ 0.086 & 20 \\
GOL-Gated CN-Coverage & 2000 & 0.350 & 0.354 $\pm$ 0.095 & 20 \\
GOL-Gated CN-Coverage & 2000 & 0.700 & 0.355 $\pm$ 0.092 & 20 \\
CN-Coverage & 200 & 0.170 & 0.359 $\pm$ 0.067 & 20 \\
CN-Coverage & 200 & 0.350 & 0.389 $\pm$ 0.071 & 20 \\
CN-Coverage & 200 & 0.700 & 0.361 $\pm$ 0.079 & 20 \\
CN-Coverage & 2000 & 0.170 & 0.397 $\pm$ 0.111 & 20 \\
CN-Coverage & 2000 & 0.350 & 0.396 $\pm$ 0.112 & 20 \\
CN-Coverage & 2000 & 0.700 & 0.391 $\pm$ 0.102 & 20 \\
\bottomrule
\end{tabular}
\end{table}

\begin{table}[H]
\centering
\caption{Geometry proxy correlation.}
\label{tab:geometry_proxy_corr}
\begin{tabular}{ll}
\toprule
Metric & Value \\
\midrule
Spearman correlation & 0.221 \\
Pearson correlation & 0.028 \\
Scenes & 20 \\
\bottomrule
\end{tabular}
\end{table}

These tables establish that the guarded-vs-naive qualitative story is not visibly tied to one voxel or $\sigma$ value, and that geometry-aware quality has directional but noisy agreement with performance gaps. They do not establish a strong linear quality predictor, nor do they provide the promised distance-metric sensitivity: a separate $\lambda_\psi$ or translation-only distance sweep is not present in the available result set and is therefore not claimed as evidence here.

\paragraph{Compute and additional baseline controls.}
We run two lightweight controls. Table~\ref{tab:exposure_control} reports a compute control where steps increase with rendered-view budget; this is a dedicated rerun and should not be read as a typo or duplicate of the main-paper scaling table. Table~\ref{tab:stochastic_baseline} adds a stochastic-greedy selection baseline at $N=\{200,2000\}$ under the same step-matched 400-step protocol.
\begin{table}[H]
\centering
\scriptsize
\setlength{\tabcolsep}{2pt}
\caption{Exposure-matched control.}
\label{tab:exposure_control}
\begin{tabular}{p{0.29\linewidth}crlrc}
\toprule
Method & N & Steps & Regime & AbsRel & Scenes \\
\midrule
Random & 0 & 400 & Step-matched & 0.357 $\pm$ 0.048 & 20 \\
Random & 200 & 400 & Step-matched & 0.382 $\pm$ 0.069 & 20 \\
Random & 200 & 437 & Exp-match & 0.397 $\pm$ 0.083 & 20 \\
Random & 2000 & 400 & Step-matched & 0.355 $\pm$ 0.080 & 20 \\
Random & 2000 & 770 & Exp-match & 0.345 $\pm$ 0.076 & 20 \\
GOL-Gated CN-Coverage & 0 & 400 & Step-matched & 0.312 $\pm$ 0.069 & 20 \\
GOL-Gated CN-Coverage & 200 & 400 & Step-matched & 0.308 $\pm$ 0.036 & 20 \\
GOL-Gated CN-Coverage & 200 & 437 & Exp-match & 0.300 $\pm$ 0.050 & 20 \\
GOL-Gated CN-Coverage & 2000 & 400 & Step-matched & 0.323 $\pm$ 0.065 & 20 \\
GOL-Gated CN-Coverage & 2000 & 770 & Exp-match & 0.325 $\pm$ 0.067 & 20 \\
\bottomrule
\end{tabular}
\end{table}

\begin{table}[H]
\centering
\footnotesize
\caption{Stochastic-greedy baseline.}
\label{tab:stochastic_baseline}
\begin{tabular}{rllr}
\toprule
N & Method & AbsRel & Scenes \\
\midrule
200 & Coverage & 0.389 $\pm$ 0.058 & 20 \\
200 & CN-Coverage & 0.399 $\pm$ 0.120 & 20 \\
200 & GOL-Gated CN-Coverage & 0.308 $\pm$ 0.036 & 20 \\
200 & StochGreedy-Cov & 0.395 $\pm$ 0.057 & 20 \\
2000 & Coverage & 0.381 $\pm$ 0.095 & 20 \\
2000 & CN-Coverage & 0.355 $\pm$ 0.090 & 20 \\
2000 & GOL-Gated CN-Coverage & 0.323 $\pm$ 0.065 & 20 \\
2000 & StochGreedy-Cov & 0.375 $\pm$ 0.096 & 20 \\
\bottomrule
\end{tabular}
\end{table}

These controls establish that more compute can improve some settings and that a stronger stochastic-greedy baseline does not overturn the main stability claim. They do not justify the stronger statement that all methods improve uniformly with extra optimization; some rows remain flat or worse in these control runs.

\appendixsection{Additional Figures}
This section provides visual diagnostics that complement the main-paper tables. Each figure is tied to a specific claim about stability, outliers, or downstream trade-offs.

\paragraph{Paired distribution visualization.}
Figure~\ref{fig:supp_scaling_delta} shows per-sequence paired $\Delta$AbsRel at $N=200$ and $N=2000$. It complements Table~\ref{tab:scaling_significance}. ``Best non-guardrail'' follows the same per-sequence oracle definition above.
This figure establishes that the guarded variant behaves like variance reduction or worst-case protection rather than a universal mean booster: heavy negative tails coexist with positive medians against some comparators. It does not establish that GOL-Gated CN-Coverage wins on every scene or against every strong baseline.

\begin{figure}[!htbp]
    \centering
    \includegraphics[width=0.80\linewidth]{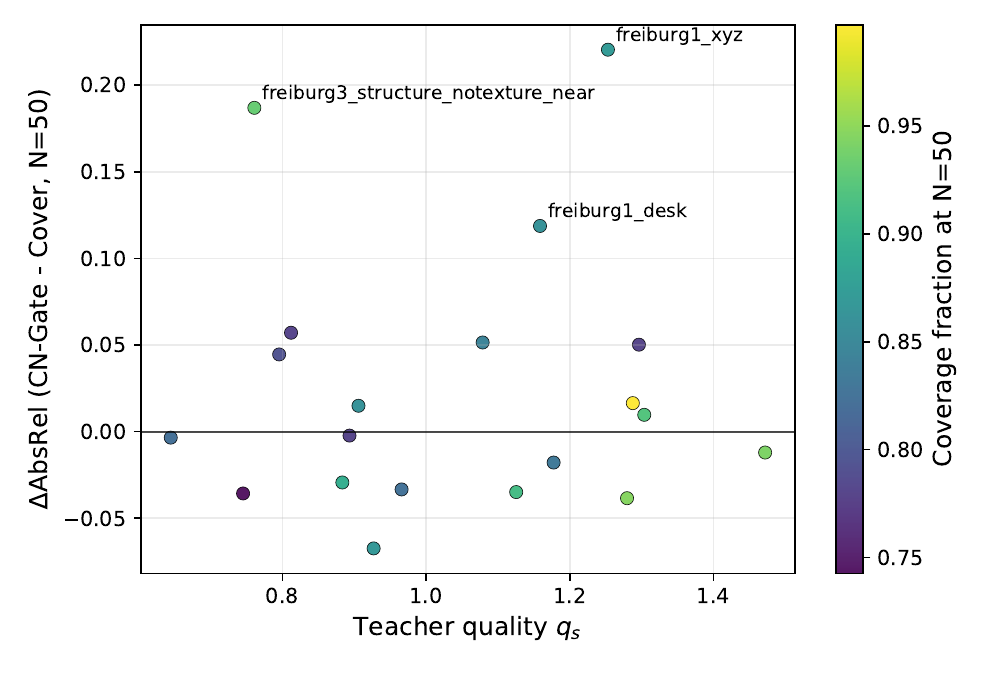}
    \caption{N=50 failure-regime diagnosis: per-sequence $\Delta$AbsRel versus teacher quality.}
    \label{fig:supp_n50_outlier}
\end{figure}

\begin{figure}[!htbp]
    \centering
    \includegraphics[width=0.80\linewidth]{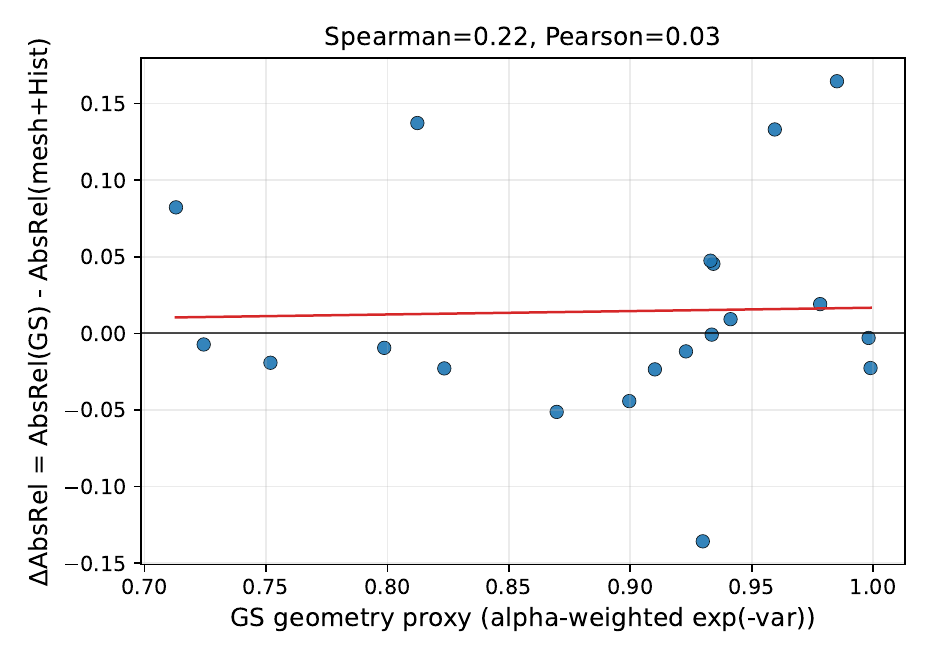}
    \caption{Geometry-aware teacher-quality proxy correlation at $N=200$.}
    \label{fig:supp_geometry_proxy}
\end{figure}

\paragraph{Outlier and teacher-quality diagnostics.}
Figure~\ref{fig:supp_n50_outlier} visualizes the $N=50$ outlier regime. Figure~\ref{fig:supp_geometry_proxy} shows that geometry-aware teacher quality has directional agreement with performance gaps, but with residual variance consistent with scene-specific effects.

These figures establish that the observed outliers cluster in a few sequences and that teacher quality is better interpreted as coarse risk control than as a precise predictor. They do not establish a calibrated or linear mapping from quality proxies to depth improvement.

\paragraph{Downstream scaling curves.}
Main Table~7 reports the shared downstream numeric summary. Figure~\ref{fig:supp_downstream_scaling} reuses the same run artifacts and adds the full per-budget curves, including the shared $N=0$ checkpoint for Coverage/CN-Coverage.
\vspace*{\fill}
\begin{strip}
    \centering
    \includegraphics[width=0.78\textwidth]{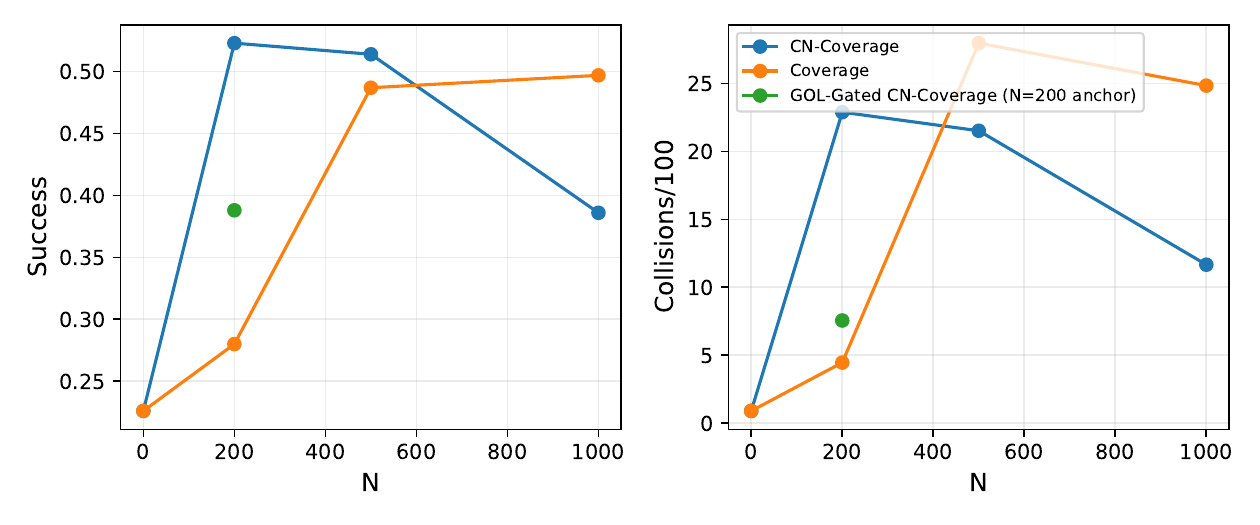}
    \captionof{figure}{Downstream control-proxy metrics versus budget $N$ for Coverage and CN-Coverage.}
    \label{fig:supp_downstream_scaling}
\end{strip}
These artifacts establish that scaling policy changes the behavior of the stylized control proxy, including a shared $N=0$ baseline for Coverage/CN-Coverage. They do not establish policy learning, policy transfer, or real-robot safety; the downstream evidence remains a controlled diagnostic under fixed kinematics and thresholds.

\appendixsection{Additional Paired Summaries}
This section adds compact summaries derived from the available per-scene logs. They complement the main paper by making paired comparisons, tail behavior, and the meaning of the teacher-quality threshold easier to inspect.

\paragraph{Direct guardrail comparisons.}
Table~\ref{tab:guardrail_pairwise} summarizes direct scene-level comparisons for GOL-Gated CN-Coverage at $N=\{200,2000\}$. $\Delta$AbsRel is target minus comparator, so negative is better, and W/L/T counts scene-wise wins, losses, and ties.
\begin{table}[H]
\centering
\footnotesize
\setlength{\tabcolsep}{3pt}
\caption{Guardrail comparisons.}
\label{tab:guardrail_pairwise}
\resizebox{\columnwidth}{!}{%
\begin{tabular}{rllrll}
\toprule
N & Comparator & MeanD & MedD & W/L/T & p \\
\midrule
200 & CN-Coverage & -0.091 & -0.010 & 11/9/0 & 0.368 \\
200 & CN-Coverage (MeshHist) & -0.076 & 0.030 & 9/11/0 & 0.869 \\
2000 & CN-Coverage & -0.032 & -0.003 & 11/9/0 & 0.498 \\
2000 & CN-Coverage (MeshHist) & -0.067 & -0.009 & 12/8/0 & 0.097 \\
\bottomrule
\end{tabular}
}
\end{table}

Table~\ref{tab:guardrail_pairwise} shows lower mean AbsRel than both direct comparators, but the scene-wise wins and losses remain close. The evidence therefore supports stability-oriented guarding rather than a universal mean-improvement claim.

\paragraph{Highest-novelty paired summary.}
Table~\ref{tab:tail_pairwise} reports paired comparisons in the most novel per-sequence quantile bin (5/5) at $N=0$ and $N=2000$. Here too, $\Delta$AbsRel is target minus comparator, so negative is better.
\begin{table}[H]
\centering
\footnotesize
\setlength{\tabcolsep}{3pt}
\caption{Tail comparisons.}
\label{tab:tail_pairwise}
\resizebox{\columnwidth}{!}{%
\begin{tabular}{rllrll}
\toprule
N & Comparator & MeanD & MedD & W/L/T & p \\
\midrule
0 & CN-Coverage & -0.048 & -0.066 & 17/3/0 & 0.005 \\
0 & CN-Coverage (MeshHist) & -0.011 & 0.003 & 10/10/0 & 0.674 \\
2000 & CN-Coverage & -0.039 & -0.005 & 12/8/0 & 0.216 \\
2000 & CN-Coverage (MeshHist) & -0.060 & -0.016 & 15/5/0 & 0.012 \\
\bottomrule
\end{tabular}
}
\end{table}

Table~\ref{tab:tail_pairwise} establishes that the guarded variant has the strongest paired evidence in the highest novelty bin against CN-Coverage (MeshHist), especially at $N=2000$. It does not establish superiority over every comparator in every novelty regime, which is why the main paper keeps the tail claim narrowly focused on the guarded variant.

\paragraph{Teacher-quality threshold rationale.}
\vspace*{\fill}
\begin{strip}
    \centering
    \includegraphics[width=0.72\textwidth]{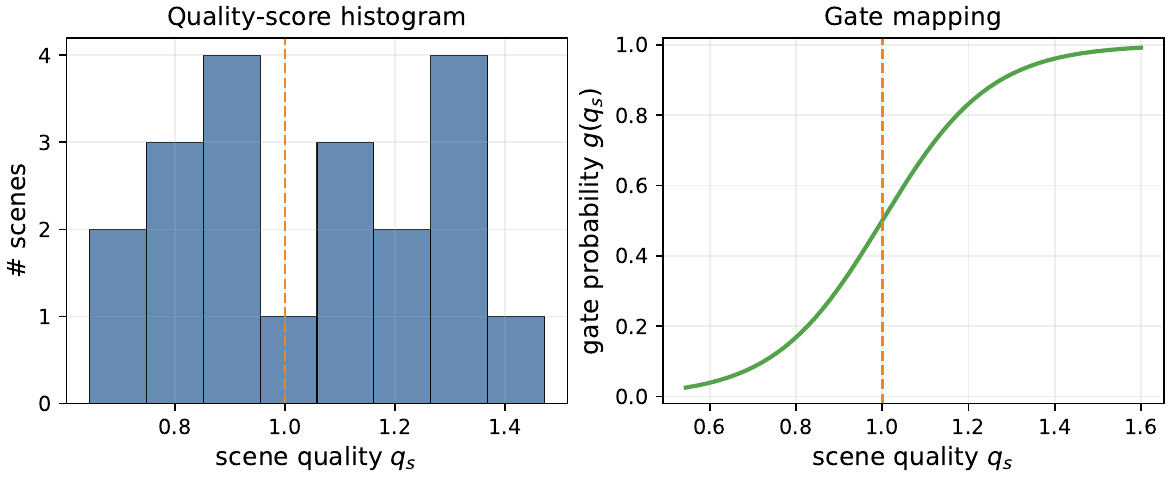}
    \captionof{figure}{Scene-quality threshold view. Left: histogram of scene-level quality score $q_s$ over the 20 TUM sequences. Right: gate mapping $g(q_s)=\operatorname{sigmoid}(8(q_s-1))$ used by GOL-Gated. The threshold $q_s=1$ is the point at which all three held-out val-RGB criteria meet their nominal bounds.}
    \label{fig:qs_threshold}
\end{strip}
Figure~\ref{fig:qs_threshold} establishes the practical meaning of the gate threshold and shows that the benchmark splits evenly into high- and low-quality scene buckets. It does not establish that the threshold is uniquely optimal; the available evidence only supports it as a reasonable coarse risk-control rule.

\end{document}